\documentclass[journal]{IEEEtran}

\usepackage{cite}
\usepackage{graphicx}
\usepackage{amssymb,amsmath,bm}

\usepackage{multicol}
\usepackage{multirow}
\usepackage{textcomp}
\usepackage{balance}
\usepackage{subcaption}
\usepackage{xcolor}
\usepackage{pgf}
\usepackage{threeparttable}
\usepackage{booktabs}
\usepackage{tikz}
\usetikzlibrary{automata,arrows,positioning,chains,shapes.multipart,fit,petri,shapes,backgrounds,decorations.pathmorphing,calc,shapes.misc, arrows, decorations.markings}
\usepackage{ifthen}

\newcommand{\eg}{e.\,g.,\ }
\newcommand{\ie}{i.\,e.,\ }

\newcommand{\et}{{et al.}}

\usepackage{soul}

\ifCLASSINFOpdf
\else
\fi

\begin{document}
\title{Adversarial Training in Affective Computing\\and Sentiment Analysis: \\Recent Advances and Perspectives
}

\author{
Jing~Han, %~\IEEEmembership{Student~Member,~IEEE,}
Zixing~Zhang, 
Nicholas~Cummins, 
and~Bj\"orn~Schuller% <-this % stops a space
% Jing~Han, %~\IEEEmembership{Student~Member,~IEEE,}
% Zixing~Zhang,~\IEEEmembership{Member,~IEEE,}
% Nicholas~Cummins,~\IEEEmembership{Member,~IEEE,} 
% and~Bj\"orn~Schuller,~\IEEEmembership{Fellow,~IEEE}% <-this % stops a space
\thanks{J. Han and N. Cummins are with the ZD.B Chair of Embedded Intelligence for Health Care and Wellbeing, University of Augsburg, Germany.
% (e-mail: jing.han@informatik.uni-augsburg.de).
}% <-this % stops a space
\thanks{Z. Zhang is with GLAM -- Group on Language, Audio \& Music, Imperial College London, UK (corresponding author, email: zixing.zhang@imperial.ac.uk).}% <-this % stops a space
\thanks{B. Schuller is with the ZD.B Chair of Embedded Intelligence for Health Care and Wellbeing, University of Augsburg, Germany, and also with GLAM -- Group 
on Language, Audio \& Music, Imperial College London, UK.
% (e-mail: bjoern.schuller@imperial.ac.uk).
}% <-this % stops a space
}

% \markboth{IEEE Computational Intelligence Magazine,~2018}%
% {Han \MakeLowercase{\textit{et al.}}: ******}

\maketitle

\begin{abstract}

Over the past few years, adversarial training has become an extremely active research topic and has been successfully applied to various Artificial Intelligence (AI) domains. As a potentially crucial technique for the development of the next generation of emotional AI systems, we herein provide a comprehensive overview of the application of adversarial training to affective computing and sentiment analysis. Various representative adversarial training algorithms are explained and discussed accordingly, aimed at tackling diverse challenges associated with emotional AI systems. Further, we highlight a range of potential future research directions. We expect that this overview will help facilitate the development of adversarial training for affective computing and sentiment analysis in both the academic and industrial communities.  

\end{abstract}

\IEEEpeerreviewmaketitle

%%%%%%%%%%%%%%%%%%%%%%%%%%%%%%%%%%%%%%%%%%%%%%%%%%%%%%%%%%%%%%%%%%%%%%
\section{Introduction}
\label{sec:introduction}

\noindent
Affective computing and sentiment analysis currently play a vital role in transforming current \emph{Artificial Intelligent} (AI) systems into the next generation of emotional AI devices~\cite{Picard97-AC,Minsky07-TEM}. It is a highly interdisciplinary research field spanning psychology, cognitive, and computer science. Its motivations include endowing machines with the ability to detect and understand the emotional states of humans and, in turn, respond accordingly~\cite{Picard97-AC}. Both the terms \emph{affective computing} and \emph{sentiment analysis} relate to the computational interpretation and generation of human emotion or affect. Whereas the former mainly relates to instantaneous emotional expressions and is more commonly associated with speech or image/video processing, the later mainly relates to longer-term opinions or attitudes and is more commonly associated with natural language processing.

A plethora of applications can benefit from the development of affective computing and sentiment analysis~\cite{Pantic05-AMH,Poria15-SDF,Poria15-TAI,Cambria16-ACA,Chen16-LUA,Han17-SMF}; examples include natural and friendly human--machine interaction systems, intelligent business and customer service systems, and remote health care systems. Thus, affective computing and sentiment analysis attract considerable research attention in both the academic and industrial communities.

From a technical point of view, affective computing and sentiment analysis are associated with a wide range of advancements in machine learning, especially in relation to deep learning technologies. For example, deep \emph{Convolutional Neural Networks} (CNNs) have been reported to considerably outperform conventional models and non-deep neural networks on two benchmark databases for sentiment analysis~\cite{Santos14-DCN}. Further, an end-to-end deep learning framework which automatically learns high-level representations from raw audio and video signals has been shown to be effective for emotion recognition~\cite{Tzirakis17-EME}.

However, when deployed in real-life applications, affective computing and sentiment analysis systems face many challenges. These include the sparsity and unbalance problems of the training data~\cite{Zhang17-ADE}, the instability of the emotion recognition models~\cite{Chen17-ADA,Deng18-SAF}, and the poor quality of the generated emotional samples~\cite{Rajeswar17-AGO,Gao18-VIU}. Despite promising research efforts and advances in leveraging techniques, such as semi-supervised learning and transfer learning~\cite{Zhang17-ADE}, finding robust solutions to these challenges is an open and ongoing research challenge. 

In 2014, a novel learning algorithm called {\em adversarial training} (or adversarial learning) was proposed by Goodfellow~\et~\cite{Goodfellow14-GAN}, and has attracted widespread research interests across a range of machine learning domains~\cite{Wang17-GAN,Creswell18-GAN}, including affective computing and sentiment analysis~\cite{Radford16-URL,Han18-TCA,Fedus18-MBT}. The initial adversarial training framework, Generative Adversarial Networks (GANs), consists of two neural networks -- a generator and a discriminator, which contest with each other in a two-player zero-sum game. The generator aims to capture the potential distribution of real samples and generates new samples to `cheat' the discriminator as far as possible, whereas the discriminator, often a binary classifier, distinguishes the sources (\ie real samples or generated samples) of the inputs as accurately as possible. Since its inception, adversarial training has been frequently demonstrated to be effective in improving the robustness of recognition models and the quality of the simulated samples~\cite{Goodfellow14-GAN,Wang17-GAN,Creswell18-GAN}. 

% Thus, adversarial training has emerged as an efficient tool to help overcome the aforementioned problems. Over the past three years, the number of related % papers grows exponentially. This has created a necessity to have a survey to summarise the recent research trends and directions of adversarial training in 
% affective computing and sentiment analysis.

Thus, adversarial training is emerging as an efficient tool to help overcome the aforementioned challenges when building affective computing and sentiment analysis systems. More specifically, on the one hand, GANs have the potential to produce an unlimited amount of realistic emotional samples; on the other hand, various GAN variants have been proposed to learn robust high-level representations. Both of the aspects can improve the performance of emotion recognition systems. Accordingly, over the past three years, the number of related papers has grown exponentially. Motived by the pronounced improvement achieved by these works and by the belief that adversarial training can further advance more works in the community, we thus feel that, there is a necessity to summarise recent studies, and draw attention to the emerging research trends and directions of adversarial training in affective computing and sentiment analysis.

A plethora of surveys can be found in the relevant literature either focusing on conventional approaches or (non-adversarial) deep-learning approaches for both affective computing~\cite{Zeng09-ASO, Calvo10-ADA, Gunes11-ERA, Zhang17-ADE, Schuller18-SER} and sentiment analysis~\cite{Liu12-SAA, Medhat14-SAA, Liu15-SAM,  Cambria16-ACA, Cambria17-SAI, Poria17-ARO, Soleymani17-ASO, Zhang18-DLFS}, or offering more generic overviews of generative adversarial networks~\cite{Wang17-GAN,Creswell18-GAN}. Differing from these surveys, the present article:
\begin{itemize}
\item provides, for the first time, a comprehensive overview of the adversarial training techniques developed for affective computing and sentiment analysis applications; 
\item summarises the adversarial training technologies suitable, not only for the emotion recognition and understanding tasks, but more importantly, for the 
emotion synthesis and conversion tasks, which are arguably far from being regarded as mature;  
\item reviews a wide array of adversarial training technologies covering the text, speech, image and video modalities;
\item highlights an abundance of future research directions for the application of adversarial training in affective computing and sentiment analysis.
\end{itemize}

The remainder of this article is organised as follows. In Section~\ref{sec:background}, we first introduce the background of this overview, which is then followed by a short description of adversarial training in Section~\ref{sec:adversarialTraining}. We then comprehensively summarise the representative  adversarial training approaches for emotion synthesis in Section~\ref{sec:synthesis}, the approaches for emotion conversion in Section~\ref{sec:conversion}, and the approaches for emotion perception and understanding in Section~\ref{sec:understanding}, respectively. We further highlight some promising research trends in Section~\ref{sec:futureWork}, before drawing the conclusions in Section~\ref{sec:conclusion}.

% \begin{figure}
% \caption{Number of published papers over the past three years relating to the adversarial training in affective computing and sentiment analysis.}
% \label{fig:papers}
% \end{figure}

%%%%%%%%%%%%%%%%%%%%%%%%%%%%%%%%%%%%%%%%%%%%%%%%%%%%%%%%%%%%%%%%%%%%%%
\section{Background}
\label{sec:background}
\noindent

\noindent
In this section, we first briefly describe three of the main challenges associated with affective computing and sentiment analysis, \ie the naturalness of generated emotions, the sparsity of collected data, and the robustness of trained models. Concurrently, we analyse the drawbacks and limitations of conventional deep learning approaches, and introduce opportunities for the application of adversarial training. Then, we give a short discussion about the challenge of performance evaluation when generating or converting emotional data.

\begin{figure}[t!]
  \centering
  \input{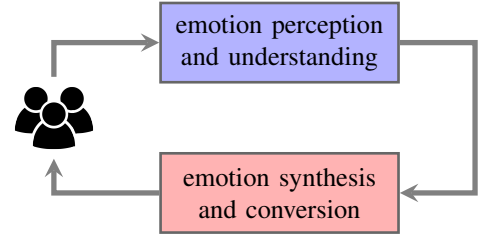}
 \caption{The broad framework of a typical emotional artificial intelligence system.}
 \label{fig:system}
\end{figure}

A typical emotional AI framework consists of two core components: an \emph{emotion perception and understanding} unit, and an \emph{emotion synthesis and conversion} unit (cf.~Figure~\ref{fig:system}). The first component ({\it aka} a recognition model) interprets human emotions; whereas the second  component ({\it aka} a generation model) can generate emotionally nuanced linguistics cues, speech, facial expressions, and even gestures. For the remainder of this article, the term \emph{emotion synthesis} refers to the artificial generation of an emotional entity from scratch, whereas the term 
\emph{emotion conversion} refers to the transformation of an entity from one emotional depiction to another. To build a robust and stable emotional AI system, several challenges have to be overcome as discussed in the following sub-sections. 

\subsection{Naturalness of Generated Emotions}
\label{subsec:naturalness}

\noindent
Emotion synthesis and conversion go beyond the conventional constructs of Natural Language Generation (NLP), Text-To-Speech (TTS), and image/video transformation techniques. This is due in part to the instinct complexity and uncertainty of the emotions, thereby generating an emotional entity remains an ongoing challenge.

Recently, research has shown the potential of deep-learning based generative models for addressing this challenge. For example, the WaveNet network developed by Oord~\et~\cite{Oord16-WAG} efficiently synthesises speech signals, and Pixel Recurrent Neural Networks (PixelRNN) and Variational AutoEncoder (VAE), proposed by Oord~\et~\cite{Oord16-Oord} and Kingma~\et~\cite{Kingma13-AEV} respectively, have been shown to be effective for generating images.

To date, the majority of these studies have not considered emotional information. A small handful of works have been undertaken in this direction~\cite{Lee17-EET, Akuzawa18-ESS}, however, the generated emotions are far from being considered natural. This is due in part to the highly non-linear nature of emotional expression changes and the variance of individuals~\cite{Qiao18-GCG,Zhou17-ECM}. Generative modelling with adversarial training, on the other hand, has frequently been shown to be powerful in regard to generating samples, which are more understandable to humans than the examples simulated by other approaches~\cite{Wang17-GAN,Creswell18-GAN} (see Section~\ref{sec:synthesis} and Section~\ref{sec:conversion} for more details). 

\subsection{Sparsity of Collected Data}
\label{subsec:sparsity}

\noindent
Despite having the possibility to collect massive amounts of unlabelled data through pervasive smart devices and social media, reliably annotated data resources required for emotion analysis are still comparatively scarce~\cite{Zhang17-ADE}. For example, most of the databases currently available for speech emotion recognition contain, at most 10\,h of labelled data~\cite{Schuller10-CCA,Schuller18-SER}, which is insufficient for building highly robust models. This issue has become even more pressing in the era of deep learning. 
The data-sparsity problem mainly lies in the annotation process which is prohibitively expensive and time-consuming~\cite{Zhang17-ADE}. This is especially true in relation to the subjective nature of emotions which dictates the need for several annotators to label the same samples in order to diminish the effect of personal biases~\cite{Han17-FHT}.

In tackling this challenge, Kim~\et~\cite{Kim13-DLF} proposed an unsupervised learning approach to learn the representations across audiovisual modalities for emotion recognition without any labelled data. Similarly, Cummins~\et~\cite{Cummins17-AID} utilised CNNs pre-trained on large amounts of image data to extract robust feature representations for speech-based emotion recognition. More recently, neural-network-based semi-supervised learning has been introduced to leverage large-scale unlabelled data~\cite{Deng18-SAF,Zhang18-LUD}. 

Despite the effectiveness of such approaches that distil shared high-level representations between labelled and unlabelled samples, the limited number of labelled data samples means that there is a lack of sufficient resources to extract meaningful and salient representations specific to emotions. In contrast, a generative model with adversarial training has the potential to synthesise an infinite amount of labelled samples to overcome the shortage of conventional deep learning approaches (see Section~\ref{sec:understanding} for more details). 

\subsection{Robustness of Trained Models}
\label{subsec:mismatch}

\noindent
In many scenarios, samples from a target domain are not sufficient or reliable enough to train a robust emotion recognition model. This challenge has motivated researchers to explore \emph{transfer learning} solutions which leverage related domain (source) samples to aid the target emotion recognition task. This is a highly non-trivial task, the source and target domains are often highly mismatched with respect to the domains  in which the data are collected~\cite{Pan10-ASO}, such as different recording environments or websites. For example, in sentiment analysis, the word `long' for evaluating battery life has a positive connotation, whereas when assessing pain it tends to be negative. Moreover, for speech emotion recognition, the source and target samples might have been recorded in distinctive acoustic environments and by different speakers~\cite{Zhang17-ADE}. These mismatches have been shown to lead to a performance degradation of models analysed in real-life settings~\cite{Pan10-ASO,Glorot11-DAF,Medhat14-SAA}.

In addressing this challenge, Glorot~\et~\cite{Glorot11-DAF} presented a deep neural network based approach to learn the robust representations across different domains for sentiment analysis. Similar approaches have also been proposed by Deng~\et~\cite{Deng14-ISA} for emotion recognition from speech. Moreover, You~\et~\cite{You15-RIS} successfully transferred the sentiment knowledge from text to predict the sentiment of images. However, it is still unclear if their learnt representations are truly domain-invariant or not. 

On the other hand, the discriminator of an adversarial training framework has the potential to  distinguish from which domain the so-called `shared' representations come from. By doing so, it can help alleviate the robustness problem of an emotion recognition model (see Section~\ref{sec:understanding} for more details). 

\subsection{Performance Evaluation}
% \nc{``Meaningful Performance Evaluation''?}

\noindent
Evaluating the performance of the generated or converted emotional samples is essential but challenging in aspects of affective computing and sentiment analysis. Currently, many of the related works directly demonstrate a few appealing samples and evaluate the performance by human judgement~\cite{Bao18-TOI,Nojavansghari18-IGA,Qiao18-GCG}. Additionally, a range of metric-based approaches have been proposed to quantitatively evaluate the adversarial training frameworks. For example, the authors in~\cite{Huang17-DGF} compared the intra-set and inter-set average Euclidean distances between different sets of the generated faces. 

Similarly, to quantitatively evaluate models for emotion conversion, other evaluation measurements raised in the literature include BiLingual Evaluation Understudy (BLEU)~\cite{Papineni02-BLEU} and Recall-Oriented Understudy for Gisting Evaluation (ROUGE)~\cite{lin2004rouge} for text, and a signal-to-noise ratio test for speech~\cite{Gao18-VIU}. However, the quantitative performance evaluation for emotion perception and understanding is more straightforward. In general, the improvement by implementing adversarial training can be reported using evaluation metrics such as unweighted accuracy, unweighted average recall, and concordance correlation coefficient~\cite{Chen17-ADA, Deng17-SDO, Mohammed18-DAF}.

\section{Principle of Adversarial Training}
\label{sec:adversarialTraining}

\noindent
In this section, we introduce the basic concepts of adversarial training, so that the interested reader can better understand the design and selection of adversarial networks for a specific task in affective computing and sentiment analysis.

\subsection{Terminology and Notation}

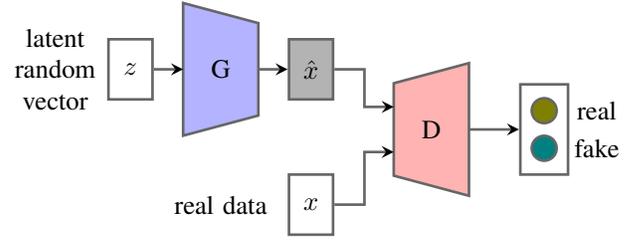
\begin{figure}[t!]
  \centering
  % \documentclass{article}
% 
% \PassOptionsToPackage{dvipsnames}{xcolor}
% \usepackage{tikz}
% \usetikzlibrary{arrows,positioning,fit,petri,backgrounds,decorations.pathmorphing,calc,shapes,shapes.misc, arrows, decorations.markings}
% 
% \usepackage[active,tightpage]{preview}
% \setlength\PreviewBorder{2pt}
% \PreviewEnvironment{tikzpicture}
% 
% \usepackage[latin1]{inputenc}
% 
% \usepackage{amsmath}
% \usepackage{amssymb}
% \newcommand{\mb}{\mathbf}
% \newcommand{\mc}{\mathcal}
% 
% \begin{document}
% \pagestyle{empty}

   \def\layersep{3cm}
   \def\hsep{1cm}
   \begin{tikzpicture}[shorten >=0pt,->,>=stealth, draw=black!60, node distance=\layersep, line width=1pt]
  
      \tikzstyle{object} = [rectangle, draw, fill=white, minimum height=3em, minimum width=5em, inner sep=0pt, align=center];
      \tikzstyle{action} = [circle, draw, text width=0.3cm,font=\small, inner sep=0pt,text centered];
      \tikzstyle{annot} = [rectangle, draw, text width=1em, minimum height=8mm, text centered];
      \tikzstyle{net} = [trapezium, fill=gray!20!white, trapezium angle=75, draw, inner xsep=0pt, outer sep=0pt, minimum height=10mm, text width=6mm,text centered];
      \tikzstyle{dot} = [draw,circle, scale=1];
      \tikzstyle{symbol} = [text centered, text width=1.4cm]; 
      
      \node[annot, text width=1em] (start) at (-0.5,0) {${z}$};
      \node[symbol, left of=start, node distance=1*\hsep] {latent random vector};  
      \node[net, right of=start, rotate=-90, node distance=1.2*\hsep, fill=blue!30!white] (lstm1) {\rotatebox[origin=c]{90}{G}};
       \node[annot, right of=lstm1, node distance=1.2*\hsep,fill=black!30!white] (pred) {$\hat{{x}}$};
       \node[annot, below of=pred, node distance=0.6*\layersep] (label) {${x}$};
       \node[symbol, left of=label, node distance=1.2*\hsep] {real data}; 
      \node[net, rotate=90, fill=red!30!white] (lstm2) at (3.5,-0.8) {\rotatebox[origin=c]{270}{D}};
      \node[annot, right of=lstm2, node distance=1.5*\hsep,text width=0.4cm, minimum height=1.2cm] (end) {};
      \node[dot, above of=end, node distance=0.25*\hsep, fill=red!50!green] (real) {}; 
     \node[dot, below of=end, node distance=0.25*\hsep, fill=blue!50!green] (fake) {};
     \node[right of=real, node distance=0.7*\hsep](real_text){real};
     \node[right of=fake, node distance=0.7*\hsep](fake_text){fake};

      \path (start) edge (lstm1) 
      (lstm1) edge (pred)
      (lstm2) edge (end); 
      \draw [->] (pred.east)  to ($(pred.east)+(0.4cm,0cm)$) to ($(pred.east)+(0.4cm,-0.5cm)$) to ($(lstm2.north)+(0cm,0.3cm)$);
       \draw [->] (label.east)  to ($(label.east)+(0.4cm,0cm)$) to ($(label.east)+(0.4cm,0.7cm)$) to ($(lstm2.north)+(0cm,-0.3cm)$);

   \end{tikzpicture}

% \end{document}
 \caption{Framework of Generative Adversarial Network (GAN).}
 \label{fig:GAN}
\end{figure}

\noindent
With the aim of generating realistic `fake' samples from a complex and high-dimensional true data distribution, the `classical' GAN, consists of two deep neural nets (as two players in a game): a \textit{generator (denoted as $G$)} and a \textit{discriminator (denoted as $D$)} (cf.~Figure~\ref{fig:GAN}). During this two-player game, the generator tries to turn input noises from a simple distribution into realistic samples to fool the discriminator, while the discriminator tries to distinguish between \textit{true} (or `\textit{real}') and \textit{generated} (or `\textit{fake}') data. 

Normally, $G$ and $D$ are trained jointly in a minimax fashion. Mathematically, the minimax objective function can be formulated as:
\begin{equation}
\label{eq:1}
\begin{aligned} 
\min_{{\theta_g}}\max_{\theta_d}V(D,G) =  &\mathbb{E}_{x\sim p_{\text{data}}(x)}[\log D_{\theta_d}(x)] + \\
& \mathbb{E}_{z\sim p_z (z)}[\log (1 - D_{\theta_d}(G_{\theta_g}(z)))] ,
\end{aligned}
\end{equation}
where $\theta_g$ and $\theta_d$ denote the parameters of $G$ and $D$, respectively; $x$ is a real data instance following the true data distribution $p_{\text{data}}(x)$; whilst $z$ is a vector randomly sampled following a simple distribution (\eg Gaussian); $G_{\theta_g}(z)$ denotes a generated data given $z$ as the input; and $D_{\theta_d}(\cdot)$ outputs the likelihood of real data given either $x$ or $G_{\theta_g}(z)$ as the input. Note that, the likelihood is in the range of (0,1), indicating to what extent the input is probably a real data instance. Consequently, during training, $\theta_g$ is updated to minimise the objective function such that $D_{\theta_d}(G_{\theta_g}(z))$ is close to 1; conversely, $\theta_d$ is optimised to maximise the objective such that $\log D_{\theta_d}(x)$ is close to 1 and $D_{\theta_d}(G_{\theta_g}(z))$ is close to 0. In other words, $G$ and $D$ are trying to optimise a different and opposing objective function, thus pushing against each other in a zero-sum game. Hence, the strategy is named as adversarial training.

Generally, the training of $G$ and $D$ is done in an iterative manner, \ie the  corresponding neural weights $\theta_d, \theta_g$ are updated in turns. Once 
training is completed, the generator $G$ is able to generate more realistic samples, while the discriminator $D$ can distinguish authentic data from fake 
data. More details of the basic GAN training process can be found in~\cite{Goodfellow14-GAN}. 

\subsection{Category of Adversarial Networks}
\noindent
Since the first GAN paradigm was introduced in 2014, numerous variants of the original GAN have been proposed and successfully exploited in many real-life applications. It is roughly estimated that to date, more than 350 variants of GANs have been presented in the literature over the last four years\footnote{https://github.com/hindupuravinash/the-gan-zoo/blob/master/gans.tsv}, infiltrating into various domains including image, music, speech, and text. For a comprehensive list and other resources of all currently named GANs, interested readers are referred to~\cite{Caesar17-gan, Hindupur18-gan}. Herein, we group these variants into four main categories: \emph{optimisation-based}, \emph{structure-based}, \emph{network-type-based}, and \emph{task-oriented.} 

{\bf Optimisation-based:} GANs in this category aim to optimise the minimax objective function to improve the stability and the speed of the adversarial training process. For instance, in the original GAN, the Jensen-Shannon (JS) divergence of the objective function can be a constant, particularly at the start of the training procedure where there is no overlap between the sampled real data and the generated data. To smooth the training of GANs, the Wasserstein GAN (WGAN) has been proposed by replacing the JS divergence with the earth-mover distance to evaluate the distribution distance between the real and generated data~\cite{Arjovsky17-WG}. 

Other GAN variants in this direction include the Energy-Based GAN (EBGAN)~\cite{zhao2016energy}, the Least Squares GAN (LSGAN)~\cite{mao2017least}, the Loss-Sensitive GAN (LS-GAN)~\cite{qi2017loss}, the Correlational GAN (CorrGAN)~\cite{patel2018correlated}, and the Mode Regularized GAN (MDGAN)~\cite{che2016mode}, to name but a few. 

{\bf Structure-based:} these GAN variants have been proposed and developed to improve the structure of conventional GAN.
For example, the conditional GAN (cGAN) adds auxiliary information to both the generator and discriminator to control the modes of the data being generated~\cite{Mirza14-CGA}, while the semi-supervised cGAN (sc-GAN) exploits the labels of real data to guide the learning procedure~\cite{odena2016semi}.
Other GAN variants in this category include the BiGAN~\cite{donahue2016adversarial}, the CycleGAN~\cite{Zhu17-UIT}, the DiscoGAN~\cite{pmlr-v70-kim17a}, the InfoGAN~\cite{Chen16-IIR}, and the Triple-GAN~\cite{chongxuan2017triple}.

{\bf Network-type-based:} in addition, several GAN variants have been named after the network topology used in the GAN configuration, such as the DCGAN based on deep convolutional neural networks~\cite{Radford16-URL}, the AEGAN based on autoencoders~\cite{luo2017learning}, the C-RNN-GAN based on continuous recurrent neural networks~\cite{mogren2016c}, the AttnGAN based on attention mechanisms~\cite{xu2017attngan}, and the CapsuleGAN based on capsule 
networks~\cite{jaiswal2018capsulegan}. 

{\bf Task-oriented:} lastly, there are also a large number of GAN variants that have been designed for a given task, thus serve their own specific interests. Examples, to name just a few, include the Sketch-GAN proposed for sketch retrieval~\cite{creswell2016adversarial}, the ArtGAN for artwork synthesis~\cite{Tan17-AAS}, the SEGAN for speech enhancement~\cite{pascual2017segan}, the WaveGAN for raw audio synthesis~\cite{donahue2018synthesizing}, and the VoiceGAN for voice impersonation~\cite{Gao18-VIU}.

%%%%%%%%%%%%%%%%%%%%%%%%%%%%%%%%%%%%%%%%%%%%%%%%%%%%%%%%%%%%%%%%%%%%%%
\section{Emotion Synthesis} 
\label{sec:synthesis}
% art creation with emotion 
% motion with emotion 

\noindent
As discussed in Section~\ref{subsec:naturalness}, the most promising generative models, for synthesis, currently include PixelRNN/CNN~\cite{Oord16-Oord, Oord16-CIG}, VAE~\cite{Kingma13-AEV}, and GANs~\cite{Goodfellow14-GAN}. Works undertaken with these models highlight their potential for creating realistic emotional samples. The PixelRNN/CNN approach, for example, can explicitly estimate the likelihood of real data with a tractable density function in order to generate realistic samples. However, the generating procedure is quite slow, as it must be processed sequentially. On the other hand, VAE defines an intractable density function and optimises a lower bound of the likelihood instead, resulting in a faster generating speed compared with PixelRNN/CNN. However, it suffers from the generation of low-quality samples. 
 
In contrast to other generative models, GANs directly learn to generate new samples through a two-player game without estimating any explicit density function, and have been shown to obtain state-of-the-art performance for a range of tasks  notably in image generation~\cite{Goodfellow14-GAN,Wang17-GAN,Caswell15-EAL}. In particular, GAN-based frameworks can help generate, in theory, an infinite amount of realistic emotional data, including samples with subtle changes which depict more nuanced emotional states. 

\subsection{Conditional-GAN-based Approaches in Image/Video}

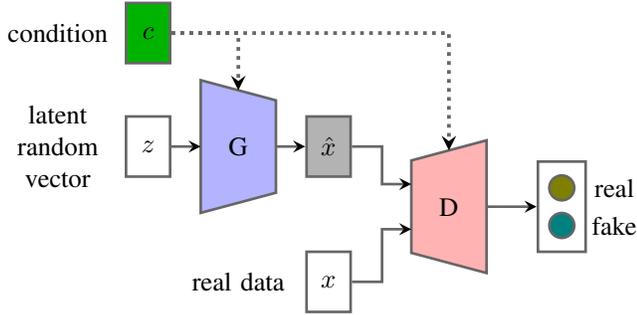
\begin{figure}[t!]
  \centering
  % \documentclass{article}
% 
% \PassOptionsToPackage{dvipsnames}{xcolor}
% \usepackage{tikz}
% \usetikzlibrary{arrows,positioning,fit,petri,backgrounds,decorations.pathmorphing,calc,shapes,shapes.misc, arrows, decorations.markings}
% 
% \usepackage[active,tightpage]{preview}
% \setlength\PreviewBorder{2pt}
% \PreviewEnvironment{tikzpicture}
% 
% \usepackage[latin1]{inputenc}
% 
% \usepackage{amsmath}
% \usepackage{amssymb}
% \newcommand{\mb}{\mathbf}
% \newcommand{\mc}{\mathcal}
% 
% \begin{document}
% \pagestyle{empty}

   \def\layersep{3cm}
   \def\hsep{1cm}
   \begin{tikzpicture}[shorten >=0pt,->,>=stealth, draw=black!60, node distance=\layersep, line width=1pt]
  
      \tikzstyle{object} = [rectangle, draw, fill=white, minimum height=3em, minimum width=5em, inner sep=0pt, align=center];
      \tikzstyle{action} = [circle, draw, text width=0.3cm,font=\small, inner sep=0pt,text centered];
      \tikzstyle{annot} = [rectangle, draw, text width=1em, minimum height=8mm, text centered];
      \tikzstyle{net} = [trapezium, fill=gray!20!white, trapezium angle=75, draw, inner xsep=0pt, outer sep=0pt, minimum height=10mm, text width=6mm,text 
centered];
      \tikzstyle{dot} = [draw,circle, scale=1];
      \tikzstyle{symbol} = [text centered, text width=1.4cm]; 
      
      \node[annot, text width=1em] (start) at (-0.5,0) {${z}$};
      \node[symbol, left of=start, node distance=1.2*\hsep] (latent) {latent random vector};  
      \node[net, right of=start, rotate=-90, node distance=1.2*\hsep, fill=blue!30!white] (lstm1) {\rotatebox[origin=c]{90}{G}};
       \node[annot, right of=lstm1, node distance=1.2*\hsep, fill=black!30!white] (pred) {$\hat{{x}}$};
       \node[annot, below of=pred, node distance=0.6*\layersep] (label) {${x}$};
       \node[symbol, left of=label, node distance=1.2*\hsep] {real data}; 
      \node[net, rotate=90, fill=red!30!white] (lstm2) at (3.5,-0.8) {\rotatebox[origin=c]{270}{D}};
%       \node[annot, right of=lstm2, node distance=1.5*\hsep,text width=0.4cm, minimum height=2cm] (end) {};
%      \node[dot, above of=end, node distance=0.7*\hsep] (dot1) {}; 
%      \node[above of=end, rotate=-90, node distance=0.2*\hsep](dots){$\cdots$};
%      \node[dot, below of=end, node distance=0.25*\hsep] (dot2) {};
%      \node[dot, below of=end, node distance=0.7*\hsep, fill=blue!50!green] (dot3) {};
%      \node[right of=dot1, node distance=1*\hsep](real_text){class-1};
%      \node[right of=dot2, node distance=1*\hsep](real_text){class-k};
%      \node[right of=dot3, node distance=1*\hsep](fake_text){fake};
      \node[annot, right of=lstm2, node distance=1.5*\hsep,text width=0.4cm, minimum height=1.2cm] (end) {};
      \node[dot, above of=end, node distance=0.25*\hsep, fill=red!50!green] (real) {}; 
     \node[dot, below of=end, node distance=0.25*\hsep, fill=blue!50!green] (fake) {};
     \node[right of=real, node distance=0.7*\hsep](real_text){real};
     \node[right of=fake, node distance=0.7*\hsep](fake_text){fake};
	  % add condition 
      \node[symbol, above of=latent, node distance=1.5\hsep] (condition) {condition};  
      \node[annot, above of=start, node distance=1.5\hsep, fill=black!30!green] (condition) {$c$};
      
      \path (start) edge (lstm1) 
      (lstm1) edge (pred)
      (lstm2) edge (end); 
      \draw [->] (pred.east)  to ($(pred.east)+(0.4cm,0cm)$) to ($(pred.east)+(0.4cm,-0.5cm)$) to ($(lstm2.north)+(0cm,0.3cm)$);
       \draw [->] (label.east)  to ($(label.east)+(0.4cm,0cm)$) to ($(label.east)+(0.4cm,0.7cm)$) to ($(lstm2.north)+(0cm,-0.3cm)$);

	   \draw[->, dotted, line width=1.4pt] (condition) -| (lstm1);
       \draw[->, dotted, line width=1.4pt] (condition) -| (lstm2);
      
   \end{tikzpicture}

% \end{document}
 \caption{Framework of conditional Generative Adversarial Network (cGAN).}
 \label{fig:CGAN}
\end{figure}

\noindent
To synthesise emotions, the most frequently used GAN relates to the conditional GAN (cGAN). In the original cGAN framework, both the generator and discriminator are conditioned on certain extra information $c$. This extra information can be any kind of auxiliary information, such as the labels or data from other modalities~\cite{Mirza14-CGA}. More specifically, the latent input noise $z$ is concatenated with the condition $c$ as a joint hidden representation for the generator $G$, in the meanwhile $c$ is combined with either the generated sample $\hat{x}$ or real data $x$ to be fed into the discriminator $D$, as demonstrated in Figure~\ref{fig:CGAN}. In this circumstance, the minimax objective function given in Equation~(\ref{eq:1}) is reformulated:
\begin{equation}
\begin{aligned} 
\label{eq:2}
\min_{{\theta_g}}\max_{\theta_d}V(D,G) =  &\mathbb{E}_{x\sim p_{\text{data}}(x)}[\log D_{\theta_d}(x\vert c)] + \\
& \mathbb{E}_{z\sim p_z (z)}[\log (1 - D_{\theta_d}(G_{\theta_g}(z\vert c)))].
\end{aligned}
\end{equation}

Recently, a collection of works have begun to explore which facial expressions and representations can best be produced via cGAN. The frameworks proposed within these works are either conditioned on attribute vectors including emotion states to generate an image for a given identity~\cite{Bao18-TOI}, or conditioned on various emotions represented by values of features such as facial action unit coefficients to produce dynamic video from a static image~\cite{Pham18-GAT}, or conditioned on arbitrary speech clips to create talking faces synchronised with the given audio sequence~\cite{Song18-TFG}. While these approaches can produce faces with convincing realism, they do not fully consider the interpersonal behaviours that are common in social interactions such as mimicry. 

In tackling this problem, one novel application was proposed in~\cite{Huang17-DGF}, in which the authors presented a cGAN-based framework to generate valid facial expressions for a virtual agent. The proposed framework consists of two stages: firstly, a person's facial expressions (in eight emotion classes) are applied as conditions to generate expressive face sketches, then, the generated sketches are leveraged as conditions to synthesise complete face images of a virtual dyad partner. However, this framework does not consider the temporal dependency on faces across various frames, can yield non-smooth facial expressions over time. In light of this, researchers in~\cite{Nojavansghari18-IGA} proposed Conditional Long Short-Term Memory networks (C-LSTMs) to synthesise contextually smooth sequences of video frames in dyadic interactions. Experimental results in~\cite{Nojavansghari18-IGA} demonstrate that the facial expressions in the generated virtual faces reflect appropriate emotional reactions to a person's behaviours.

\subsection{Other GAN-based Approaches in Image/Video}
% are only limited to... thus lacking generalisation capacity.

\noindent
In addition to the cGAN-based framework, other GAN variants such as DCGAN~\cite{Radford16-URL} and InfoGAN~\cite{Chen16-IIR} have been investigated for emotional face synthesis. In~\cite{Radford16-URL}, it is shown that, vector arithmetic operations in the input latent space can yield semantic changes to the image generations. For example, performing vector arithmetic on mean vectors \textit{``smiling woman'' - ``neutral woman'' + ``neutral man''} can create a new image with the visual concept of \textit{``smiling man''}. The InfoGAN framework, on the other hand, aims to maximise the mutual information between a small subset of the latent variables and the observation, to learn interpretable latent representations which reflect the structured semantic data distribution~\cite{Chen16-IIR}. For instance, it has been demonstrated that by varying one latent code, the emotions of the generated faces can change from stern to happy~\cite{Chen16-IIR}.

\subsection{Approaches in Other Modalities}

\noindent
As well as the generation of expressive human faces, adversarial training has also been exploited to generate emotional samples in a range of other modalities. For example, in~\cite{Melis17-TEG} modern artwork images have been automatically generated from an emotion-conditioned GAN. Interestingly, it has been observed that various features, such as colours and shapes, within the artworks are commonly correlated with the emotions which they are conditioned on. Similarly, in~\cite{Wang18-MCS} plausible motion sequences conditioned by a variety of contextual information (\eg activity, emotion), have been synthesised by a so-called sequential adversarial autoencoder. More recently,  poems conditioned by various sentiment labels (estimated from images) have been created via a multi-adversarial training approach~\cite{Liu18-BND}.

Correspondingly, adversarial training has also been investigated for both text generation~\cite{Yu17-SSG, Li17-ALF} and speech synthesis~\cite{donahue2018synthesizing}. In particular, sentence generation conditioned on sentiment (either \textit{positive} or \textit{negative}) has been conducted in~\cite{Rajeswar17-AGO} and~\cite{Fedus18-MBT}, but both only on fixed-length sequences ($11$ words in~\cite{Rajeswar17-AGO} and $40$ words in~\cite{Fedus18-MBT}). One example can be the three generated samples with a fixed length~(40 words) found in~\cite{Fedus18-MBT} (also shown in Table~\ref{fig:maskgan}). Despite the promising nature of these initial works, the performance of such networks are far off when comparing with the quality and naturalness of image generation.

% \begin{figure}[h]
% \centering
%    \includegraphics[width=.95\linewidth,clip]{maskgan.png}
%   \caption{Samples from MaskGAN on IMDB. Source:~\cite{Fedus18-MBT}}
%  \label{fig:maskgan}
% \end{figure}

\begin{table}[!t]
\caption{Generated samples on IMDB~\cite{Fedus18-MBT}}
\begin{tabular}{p{0.cm}p{7.5cm}}
&{\bf Positive}: Follow the Good Earth movie linked Vacation is a comedy that credited against the modern day era yarns which has helpful something to the modern day s best It is an interesting drama based on a story of the famed \\
&{\bf Negative}: I really can t understand what this movie falls like I was seeing it I m sorry to say that the only reason I watched it was because of the casting of the Emperor I was not expecting anything as \\
&{\bf Negative}: That s about so much time in time a film that persevered to become cast in a very good way I didn t realize that the book was made during the 70s The story was Manhattan the Allies were to \\
\end{tabular}
\label{fig:maskgan}
\end{table}

To the best of our knowledge, emotion-integrated synthesis frameworks based on adversarial training has yet to be implemented in speech. Compared with image generation, one main issue we confront in both speech and text is the varied length to generate, which, however, could also be a learnt feature in the future.
% \nc{Mention this could be done with some sort of recurrent GAN structure?}

% {\blue
% While emotion generation can result in promising results with adversarial training frameworks, evaluating the performance of the generated samples is essential 
% but quite challenging. In fact, most of the aforementioned works directly demonstrate the results (e.g, expressive faces, motion series, and emotional poetry) 
% and evaluate the performance by observation. In contrast, to quantitatively evaluate the adversarial training frameworks, the authors in~\cite{Huang17-DGF}
% compared the intra-set and inter-set average Euclidean distances between different sets of the generated faces. Similarly, mean squared error was applied as an 
% evaluation metric in~\cite{Nojavansghari18-IGA}, aiming at generating the closest sequence to ground truth.
% }

%%%%%%%%%%%%%%%%%%%%%%%%%%%%%%%%%%%%%%%%%%%%%%%%%%%%%%%%%%%%%%%%%%%%%%
\section{Emotion Conversion}
\label{sec:conversion}

\noindent
Emotion conversion is a specific style transformation task. In computer vision and speech processing domains, it targets at transforming a source emotion into a target emotion without affecting the identity properties of the subject. Whereas for NLP, sentiment transformation aims to alter the sentiment expressed in the original text while preserving its content. In conventional approaches, paired data are normally required to learn a pairwise transformation function. In this case, the data need to be perfectly time aligned to learn an effective model, which is generally achieved by time-warping.

Adversarial training, on the other hand, does away with the need to prepare the paired data as a precondition, as the emotion transformation function can be estimated in an indirect manner. In light of this, adversarial training reshapes conventional emotion conversion procedures and makes the conversion systems simpler to be implemented and used, as time-alignment is not needed. Moreover, leveraging adversarial training makes the emotion conversion procedure more robust and accurate through the associated game-theoretic approach. 

% \begin{figure}[!t]
% \centering
%    \includegraphics[width=.85\linewidth, trim={0 0 90cm 0}, clip]{exprgan.jpg}
%   \caption{Face images transformed into new images with different expression intensity levels. Source:~\cite{Ding18-EFE}}
%  \label{fig:exprgan}
% \end{figure}
\begin{figure}[!t]
\centering
   \includegraphics[width=.95\linewidth, trim={0.8cm 1.5cm 6cm 2cm}, clip]{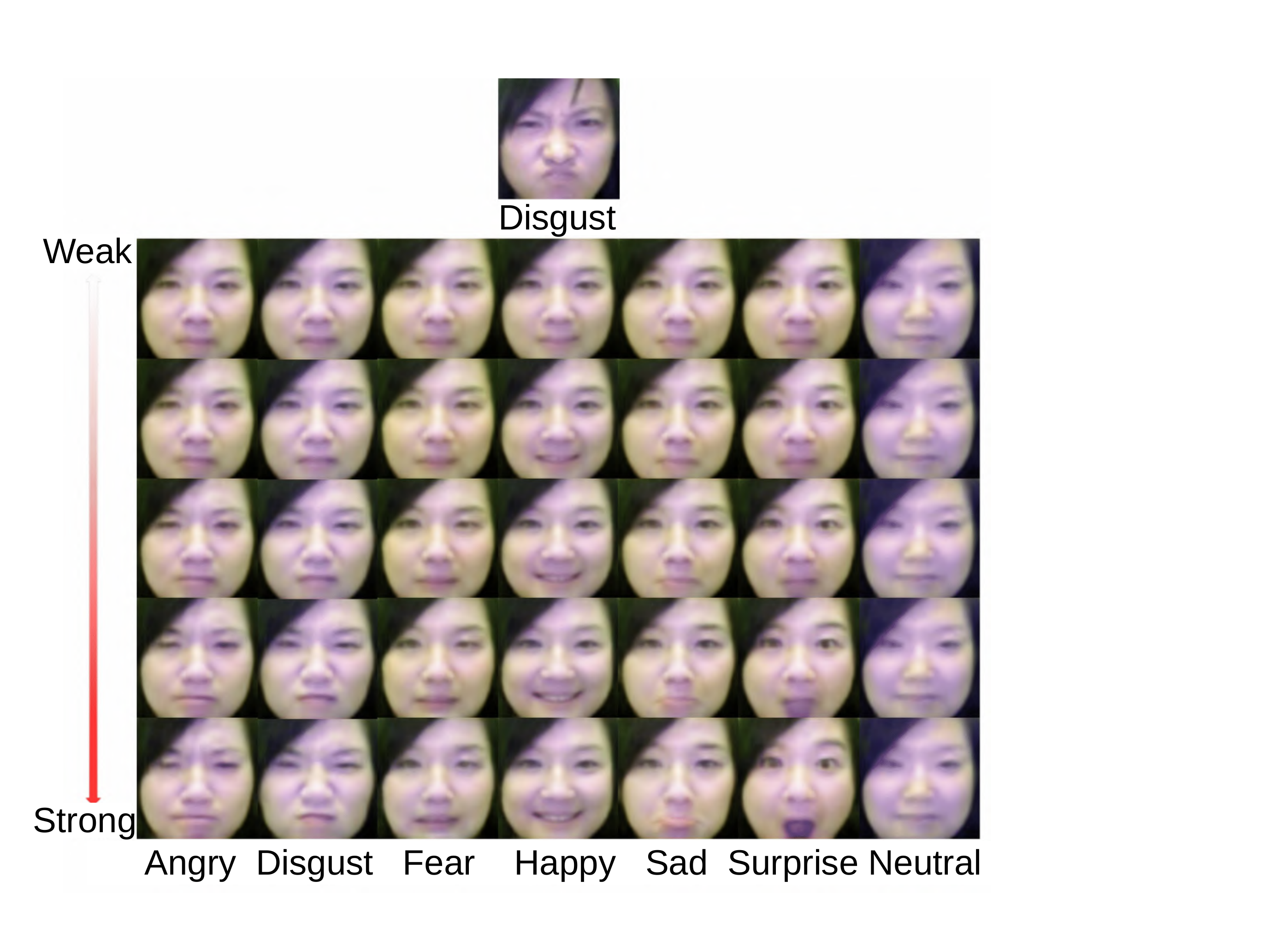}
  \caption{Face images transformed into new images with different expression intensity levels. Source:~\cite{Ding18-EFE}}
 \label{fig:exprgan}
\end{figure}

\subsection{Paired-Data-based Approaches}

\noindent
Several adversarial training approaches based on paired training data have been investigated for emotion conversion. For example, in~\cite{Zhou17-PFE}, the authors proposed a conditional difference adversarial autoencoder, to learn the difference between the source and target facial expressions of one same person. In this approach, a source face goes through an encoder to generate a latent vector representation, which is then concatenated with the target label to generate the target face through a decoder. Concurrently, two discriminators (trained simultaneously) are used to regularise the latent vector distribution and to help improve the quality of generated faces through an adversarial process.

Moreover, approaches based on facial geometry information have been proposed to guide facial expression conversion~\cite{Song17-GGA, Qiao18-GCG}. In~\cite{Song17-GGA}, a geometry guided GAN for facial expression transformation was proposed, which is conditioned on facial geometry rather than expression labels. In this way, the facial geometry is directly manipulated, and thus the network ensures a fine-grain control in face editing, which, in general, is not so straightforward in other approaches. In~\cite{Qiao18-GCG}, the researchers further disentangled the face encoding and facial geometry (in landmarks) encoding  process, which allows the model to perform the facial expression transformations appropriately even for unseen facial expression characteristics.

Another related work is~\cite{Hsu17-VCF}, in which the authors focused on voice conversion in natural speech and  proposed a variational autoencoding WGAN. Note that, data utilised in~\cite{Hsu17-VCF} are not frame aligned, but still are in pairs. Emotion conversion has not been considered in this work, however, this model could be applied to emotion conversion.

\subsection{Non-Paired-Data-based Approaches}

\noindent
The methods discussed in the previous section all require pair-wise data of the same subjects in different facial expressions during training. In contrast, Invertible conditional GAN (IcGAN), which consists of a cGAN and two encoders, does not have this constraint~\cite{Perarnau16-ICG}. In the IcGAN framework, the encoders compress a real face image into a latent representation $z$ and a conditional representation $c$ independently. Then, $c$ can be explicitly manipulated to modify the original face with deterministic complex modifications.

Additionally, the ExprGAN framework is a more recent advancement for expression transformation~\cite{Ding18-EFE}, in which the expression intensity can be controlled in a continuous manner from weak to strong. Furthermore, the identity and expression representation learning are disentangled and there is no rigid requirement of paired samples for training~\cite{Ding18-EFE}. Finally, the authors develop a three-stage incremental learning algorithm to train the model on small datasets~\cite{Ding18-EFE}. Figure~\ref{fig:exprgan} illustrates some results obtained with ExprGAN~\cite{Ding18-EFE}.

\begin{figure}[!t]
\centering
   \includegraphics[width=.85\linewidth, clip]{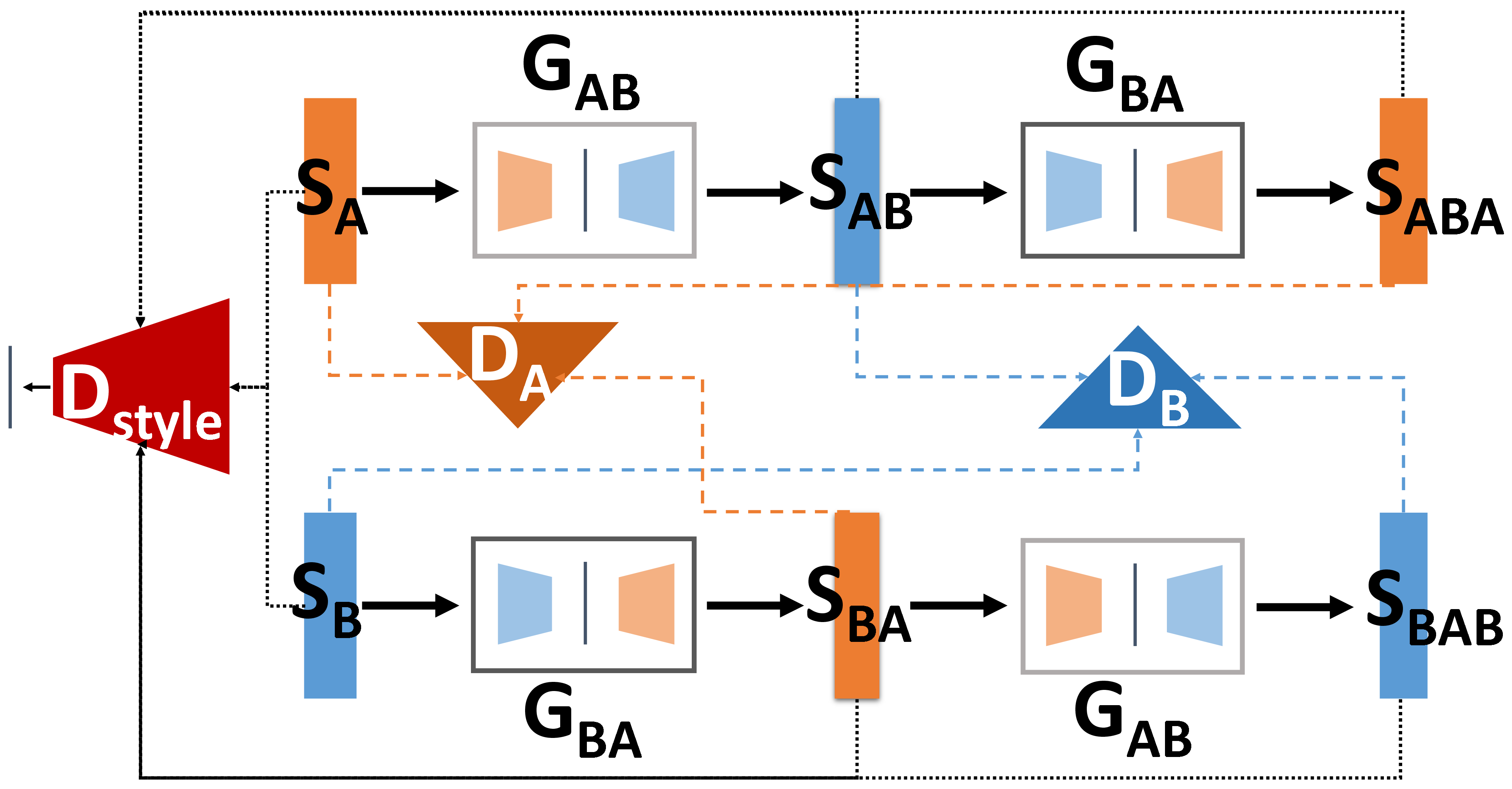}
% * <jing.han@gmx.de> 2018-09-14T15:59:46.347Z:
%
% ^.
  \caption{Framework of VoiceGAN. Source:~\cite{Gao18-VIU}}
 \label{fig:voicegan}
\end{figure}

Recently, inspired by the success of the DiscoGAN for style transformation in images, Gao \et~\cite{Gao18-VIU} proposed a speech-based style-transfer adversarial training framework, namely VoiceGAN (cf.~Figure~\ref{fig:voicegan}). The VoiceGAN framework consists of two generators/transformers ($G_{AB}$ and $G_{BA}$) and three discriminators ($D_A$, $D_B$, and $D_{style}$). Importantly, the linguistic information in the speech signals is retained by considering the reconstruction losses of the generated data, and parallel data are not required. To contend with the varied lengths of speech signals, the authors applied a channel-wise pooling to convert variable-sized feature map into a vector of fixed size~\cite{Gao18-VIU}. Experimental results demonstrate that VoiceGAN is able to transfer the gender of a speaker's voice, and this technique could be easily extended to other stylistic features such as different emotions~\cite{Gao18-VIU}.%(e.\,g.,\ different emotions)~\cite{Gao18-VIU}.

% \begin{figure}[t!]  
% \centering  
% \input{cycleGAN.tex} 
% \caption{Framework of cycle Generative Adversarial Networks (cycleGAN).} 
% \label{fig:cycleGAN}
% \end{figure}

More recently, a cycleGAN-based model was proposed to learn sentiment transformation from non-parallel text, with an ultimate goal to automatically adjust the sentiment of a chatbot response~\cite{Lee18-SSF}. By combining seq2seq model with cycleGAN, the authors developed a chatbot whose response can be transformed from negative to positive. 

Compared with the works of adversarial-training-based emotion conversion in image, it is noticeable that to date, there are only a few related works in video and speech, and only one in text.
We believe that the difficulty of applying adversarial training in these domains is threefold: 1) the variable length of corresponding sequential data; 2) the linguistic and language content needed to be maintained during the conversion; and 3) the lack of reliable measurement metrics to rapidly evaluate the performance of such a transformation.

% {\blue
% Note that, similar as in emotion generation, automatic machine evaluation other than human evaluation has been applied in order to quantitatively evaluate 
% adversarial training frameworks for emotion conversion. While samples can be converted into target emotion states, though indirect and imperfect, classification 
% test can be conducted by employing an independently-trained classifier, either to count how many generated data can be correctly classified into the target 
% class~\cite{Gao18-VIU, Ding18-EFE}, or to compute how many emotion categories the model is able to transfer~\cite{Singh18-STU}. Moreover, other evaluation 
% measurements 
% raised in the literature include BLEU~\cite{Papineni02-BLEU} and ROUGE~\cite{lin2004rouge} for text, and signal-to-noise ratio test for speech~\cite{Gao18-VIU}.
% }

%%%%%%%%%%%%%%%%%%%%%%%%%%%%%%%%%%%%%%%%%%%%%%%%%%%%%%%%%%%%%%%%%%%%%%
\section{Emotion Perception and Understanding}
\label{sec:understanding}

\noindent
This section summarises works which tackle the data-sparsity challenge (see Section~\ref{subsec:dataAug}) and the robustness-of-the-emotion-recogniser challenge (see Sections~\ref{subsec:domainAdv} and~\ref{subsec:virtualAdv}).

\subsection{Data Augmentation} 
\label{subsec:dataAug}
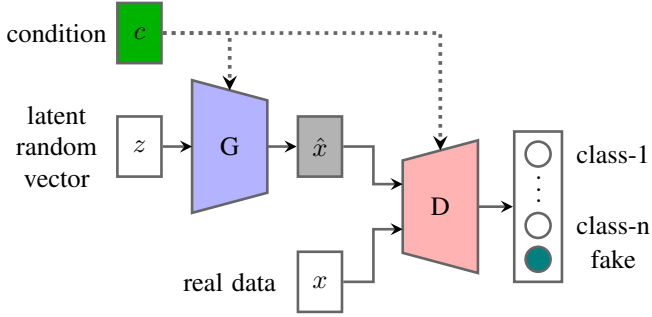
\begin{figure}[t!]
   \centering
   % \documentclass{article}
% 
% \PassOptionsToPackage{dvipsnames}{xcolor}
% \usepackage{tikz}
% \usetikzlibrary{arrows,positioning,fit,petri,backgrounds,decorations.pathmorphing,calc,shapes,shapes.misc, arrows, decorations.markings}
% 
% \usepackage[active,tightpage]{preview}
% \setlength\PreviewBorder{2pt}
% \PreviewEnvironment{tikzpicture}
% 
% \usepackage[latin1]{inputenc}
% 
% \usepackage{amsmath}
% \usepackage{amssymb}
% \newcommand{\mb}{\mathbf}
% \newcommand{\mc}{\mathcal}
% 
% \begin{document}
% \pagestyle{empty}

   \def\layersep{3cm}
   \def\hsep{1cm}
   \begin{tikzpicture}[shorten >=0pt,->,>=stealth, draw=black!60, node distance=\layersep, line width=1pt]
  
      \tikzstyle{object} = [rectangle, draw, fill=white, minimum height=3em, minimum width=5em, inner sep=0pt, align=center];
      \tikzstyle{action} = [circle, draw, text width=0.3cm,font=\small, inner sep=0pt,text centered];
      \tikzstyle{annot} = [rectangle, draw, text width=1em, minimum height=8mm, text centered];
      \tikzstyle{net} = [trapezium, fill=gray!20!white, trapezium angle=75, draw, inner xsep=0pt, outer sep=0pt, minimum height=10mm, text width=6mm,text 
centered];
      \tikzstyle{dot} = [draw,circle, scale=1];
      \tikzstyle{symbol} = [text centered, text width=1.4cm]; 
      
      \node[annot, text width=1em] (start) at (-0.5,0) {${z}$};
      \node[symbol, left of=start, node distance=1.1*\hsep] (latent) {latent random vector};  
      \node[net, right of=start, rotate=-90, node distance=1.2*\hsep, fill=blue!30!white] (lstm1) {\rotatebox[origin=c]{90}{G}};
       \node[annot, right of=lstm1, node distance=1.2*\hsep, fill=black!30!white] (pred) {$\hat{{x}}$};
       \node[annot, below of=pred, node distance=0.6*\layersep] (label) {${x}$};
       \node[symbol, left of=label, node distance=1.2*\hsep] {real data}; 
      \node[net, rotate=90, fill=red!30!white] (lstm2) at (3.5,-0.8) {\rotatebox[origin=c]{270}{D}};
      \node[annot, right of=lstm2, node distance=1.3*\hsep,text width=0.4cm, minimum height=2cm] (end) {};
     \node[dot, above of=end, node distance=0.7*\hsep] (dot1) {}; 
     \node[above of=end, rotate=-90, node distance=0.2*\hsep](dots){$\cdots$};
     \node[dot, below of=end, node distance=0.25*\hsep] (dot2) {};
     \node[dot, below of=end, node distance=0.7*\hsep, fill=blue!50!green] (dot3) {};
     \node[right of=dot1, node distance=1*\hsep](real_text){class-1};
     \node[right of=dot2, node distance=1*\hsep](real_text){class-n};
     \node[right of=dot3, node distance=1*\hsep](fake_text){fake};
%       \node[annot, right of=lstm2, node distance=1.5*\hsep,text width=0.4cm, minimum height=1.2cm] (end) {};
%       \node[dot, above of=end, node distance=0.25*\hsep, fill=red!50!green] (real) {}; 
%      \node[dot, below of=end, node distance=0.25*\hsep, fill=blue!50!green] (fake) {};
%      \node[right of=real, node distance=0.7*\hsep](real_text){real};
%      \node[right of=fake, node distance=0.7*\hsep](fake_text){fake};
	  % add condition 
      \node[symbol, above of=latent, node distance=1.5\hsep] (condition) {condition};  
      \node[annot, above of=start, node distance=1.5\hsep, fill=black!30!green] (condition) {$c$};
      
      \path (start) edge (lstm1) 
      (lstm1) edge (pred)
      (lstm2) edge (end); 
      \draw [->] (pred.east)  to ($(pred.east)+(0.4cm,0cm)$) to ($(pred.east)+(0.4cm,-0.5cm)$) to ($(lstm2.north)+(0cm,0.3cm)$);
       \draw [->] (label.east)  to ($(label.east)+(0.4cm,0cm)$) to ($(label.east)+(0.4cm,0.7cm)$) to ($(lstm2.north)+(0cm,-0.3cm)$);

	   \draw[->, dotted, line width=1.4pt] (condition) -| (lstm1);
       \draw[->, dotted, line width=1.4pt] (condition) -| (lstm2);
   \end{tikzpicture}

% \end{document}
 \caption{Framework of semi-supervised conditional Generative Adversarial Network (scGAN).}
  \label{fig:semi-GAN}
 \end{figure}

\noindent
As already discussed, the lack of large amounts of reliable training data is a major issue in the fields of affective computing and sentiment analysis. In this regard, it has been shown that emotion recognition performance can be improved with various data augmentation paradigms~\cite{Schuller12-SSF,Zhang17-AAN}. Data augmentation is a family of techniques which artificially generate more data to train a more efficient (deep) learning model for a given task. 

Conventional data augmentation methods focus on generating data through a series of transformations, such as scaling and rotating an image, or adding noise to speech~\cite{Zhang17-AAN}. However, such perturbations directly on original data are still, to some extent, not efficient to improve overall data distribution estimation. In contrast, as GANs generate realistic data which estimate the distribution of the real data, it is instinctive to apply them to expand the training data required for emotion recognition models. In this regard, some adversarial training based data augmentation frameworks have been proposed in the literature, which aim to supplement the data manifold to approximate the true distribution~\cite{Zhu17-ECW}.

For speech emotion recognition, researchers in~\cite{Sahu17-AAF} implemented an adversarial autoencoder model. In this work, high-dimensional feature vectors of real data are encoded into 2-D dimensional representations, and a discriminator is learnt to distinguish real 2-D vectors from generated 2-D vectors. The experiments indicate that the 2-D representations of real data can yield suitable margins between different emotion categories. Additionally, when adding the generated data to the original data for training, performance can be marginally increased~\cite{Sahu17-AAF}.

Similarly, a cycleGAN has been utilised for face-based emotion recognition~\cite{Zhu17-ECW}. To tackle the data inadequacy and unbalance problems, faces in different emotions have been generated from non-emotion ones, particularly for emotions like disgust and sad, which seemingly have less available samples. Experimental results have demonstrated that, by generating auxiliary data of minority classes for training, not only did the recognition performance of the rare class improve, the average performance over all classes also increased~\cite{Zhu17-ECW}. 

One ongoing research issue relating to GANs is how best to label the generated data. In~\cite{Sahu17-AAF}, they adopted a Gaussian mixture model which is built on the original data, whereas the authors in~\cite{Zhu17-ECW} took a set of class-specific GANs to generate images, respectively, which requires no additional annotation process.  

In addition to these two approaches, cGAN in a semi-supervised manner (scGAN) can be an interesting alternative worthy of future investigations. The scGAN is an extension of cGAN by forcing the discriminator $D$ to output class labels as well as distinguishing real data from fake data. In this scenario, $D$ acts as both a discriminator and a classifier. More specifically, $D$ classifies the real samples into the first $n$ classes and the generated samples into the $n+1$-th class (fake), while $G$ tries to generate the conditioned samples and `cheat' the discriminator to be correctly classified into the first $n$ classes, as illustrated in Figure~\ref{fig:semi-GAN}. By taking the class distribution into the objective function, an overall improvement in the quality of the generated samples was observed~\cite{odena2016semi}. Hence, scGAN can be easily adapted for data augmentation in emotion perception and understanding tasks, which to date has yet to be reported in the literature.

Finally, the quality of the generated data is largely overlooked in the works discussed in this section. It is possible that the generated data might be unreliable, and thus become a form of noise in the training data. In this regard, data filtering approaches should be considered.

% \subsection{Data Mismatch -- Domain Adversarial Training}
\subsection{Domain Adversarial Training}
\label{subsec:domainAdv}
% \begin{figure}[!t]
% \centering
%    \includegraphics[width=.99\linewidth, clip, height=1.6in, trim={20mm 0mm 15mm 0mm}]{dann.png}
%   \caption{Framework of Domain-Adversarial Neural Network (DANN). Source:~\cite{Ganin16-DAT}}
%  \label{fig:dann}
% \end{figure}
\begin{figure}[!t]
\centering
   \includegraphics[width=.99\linewidth, clip,trim={0mm 25mm 5mm 0mm}]{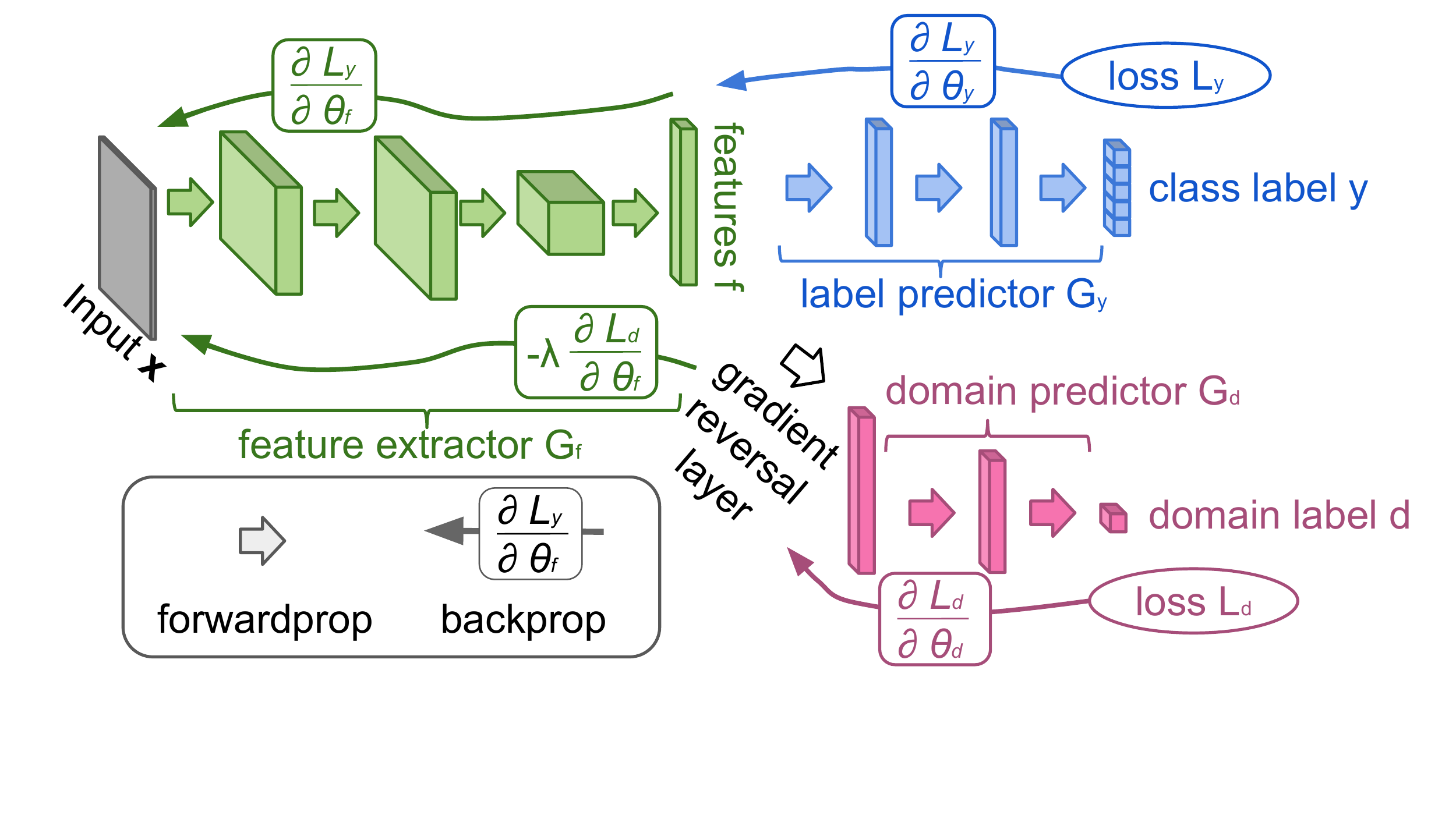}
  \caption{Framework of Domain-Adversarial Neural Network (DANN). Source:~\cite{Ganin16-DAT}}
 \label{fig:dann}
\end{figure}

\noindent
For emotion perception and understanding, numerous domain adaptation approaches have been proposed to date (cf. Section~\ref{subsec:mismatch}). These approaches seek to extract the most representative features from the mismatched data between the training phase and the test phase, in order to improve the robustness of recognition models (cf. Section~\ref{subsec:mismatch}). However, it is unclear if the learnt representations are truly domain-generative or still domain-specific.  

\begin{figure}
        \begin{subfigure}[b]{0.25\textwidth}
                \centering
                \includegraphics[width=.92\linewidth, trim={2cm 0cm 0cm 1cm}, clip]{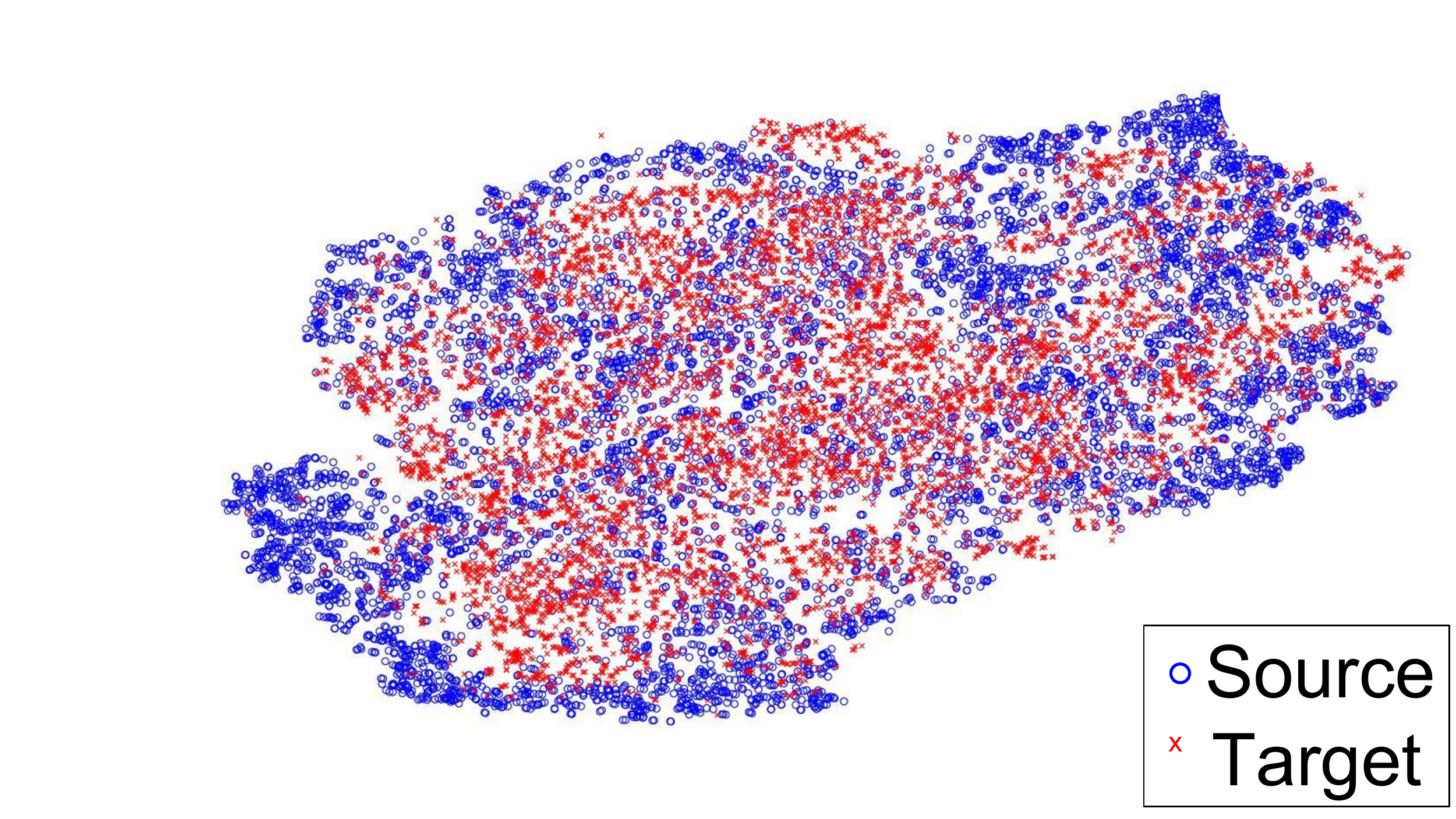}
                \caption{before DANN}
                \label{fig:gull}
        \end{subfigure}%
        \begin{subfigure}[b]{0.25\textwidth}
                \centering
                \includegraphics[width=.92\linewidth, trim={2cm 0cm 0cm 1cm}, clip]{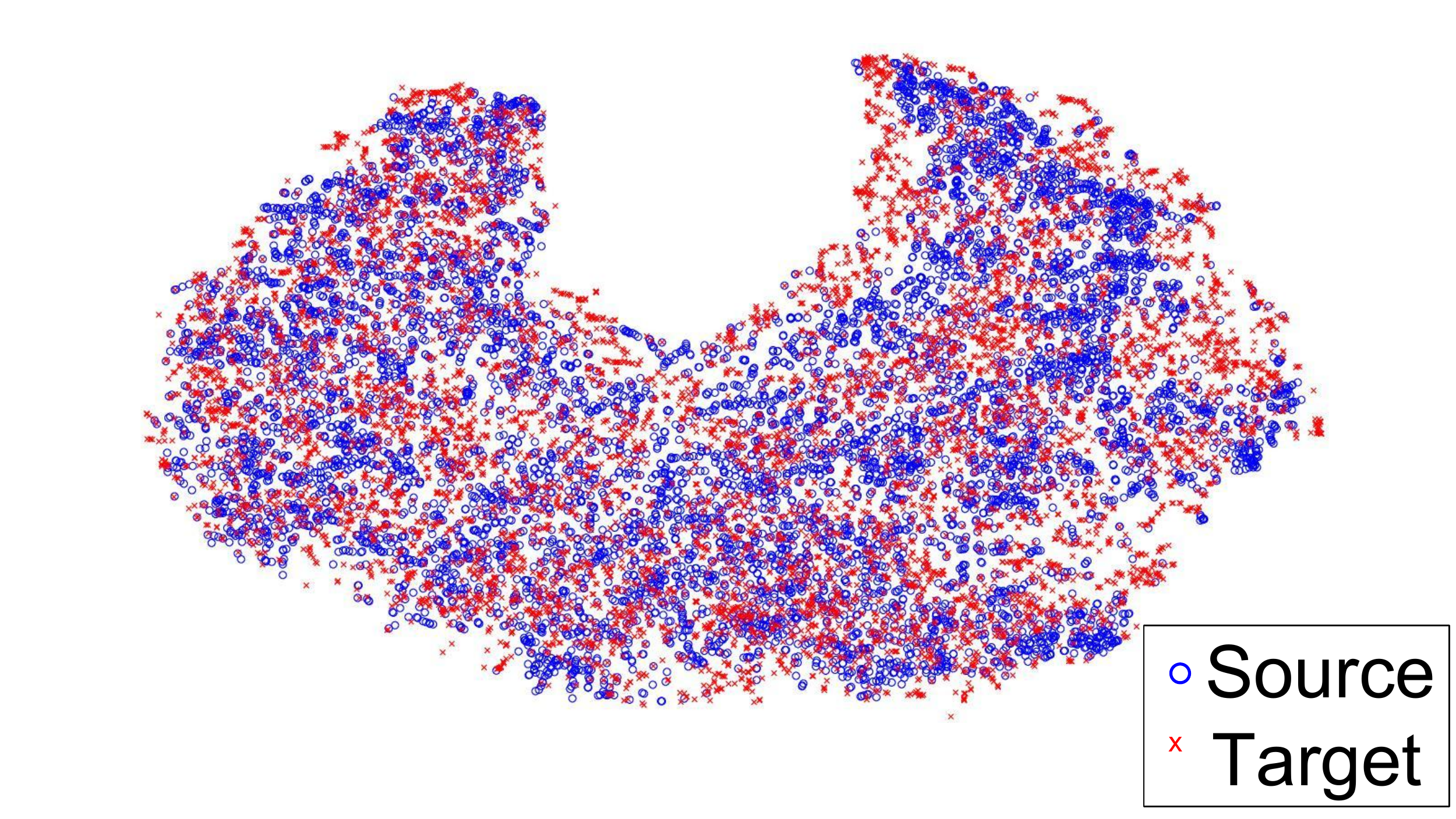}
                \caption{after DANN}
                \label{fig:gull2}
        \end{subfigure}%
        \caption{Illustration of data distributions from source and target before (a) and after (b) the DANN training. 
Source:~\cite{Mohammed18-DAF}}\label{fig:animals}
\end{figure}

In~\cite{Ganin16-DAT}, Ganin~\et~first introduced domain adversarial training to tackle this problem. Typically, a \textit{feature extractor} $G_f(:,\theta_f)$ projects data from two separate domains into high-level representations, which are discriminative for a \textit{label predictor} $G_y(:,\theta_y)$ and indistinguishable for a \textit{domain classifier} $G_d(:,\theta_d)$. A typical \emph{Domain-Adversarial Neural Network}~(DANN) is illustrated in Figure~\ref{fig:dann}. Particular to the DANN architecture, a \textit{gradient reversal layer} is introduced between the domain classifier and the feature extractor, which inverts the sign of the gradient $\frac{\partial {L_d}}{\partial {\theta_d}}$ during backward propagation.
Moreover, a hype-parameter $\lambda$ is utilised to tune the trade-off between the two branches during the learning process. In this manner, the network attempts to learn domain-invariant feature representations. By this training strategy, the representations learnt from different domains cannot be easily distinguished, as demonstrated in Figure~\ref{fig:animals}. Further details on how to train DANN are given in~\cite{Ganin16-DAT}.

% \begin{figure}[!t]
% \centering
%    \includegraphics[width=.99\linewidth, clip, height=1.6in, trim={20mm 0mm 15mm 0mm}]{dann.png}
%   \caption{Framework of Domain-Adversarial Neural Network (DANN). Source:~\cite{Ganin16-DAT}}
%  \label{fig:dann}
% \end{figure}

Using a DANN, a common representation between data from the source and target domains can potentially be learnt. This is of relevance in the data mismatch scenario as knowledge learnt from the source domain can be applied directly to the target domain~\cite{Mohammed18-DAF}. Accordingly, the original DANN paradigm has been adapted to learn domain-invariant representations for sentiment classification. For example, in~\cite{Li17-EAM, Zhang17-AAN}, attention mechanisms were introduced to give more attention to relevant text when extracting features. In~\cite{Shen18-WDG, Chen17-ADA}, the Wasserstein distance was estimated to guide the optimisation of the domain classifier. Moreover, instead of learning common representations between two domains, other research has broadened this concept to tackle the data mismatch issue among multiple probability distributions. In this regard, DANN variants have been proposed for multiple source domain adaptation~\cite{Zhao18-MSD}, multi-task learning~\cite{Liu18-AML}, and multinomial adversarial nets~\cite{Chen18-MAN}.
 
Finally, DANN has recently been utilised in speech emotion recognition\cite{Mohammed18-DAF}. These experiments demonstrate that, by aligning the data distributions of the source and target domains (illustrated in Figure~\ref{fig:animals}), an adversarial training approach can yield a large performance gain in the target domain~\cite{Mohammed18-DAF}.

%     \begin{figure}
%         \begin{subfigure}[b]{0.50\textwidth}
%                 \centering
%                 \includegraphics[width=.65\linewidth, trim={1.1cm 1.1cm 0cm 0}, clip]{Act1_Iemocap_src.png}
%                 \caption{before DANN}
%                 \label{fig:gull}
%         \end{subfigure}%
%         \vskip\baselineskip
%         \begin{subfigure}[b]{0.50\textwidth}
%                 \centering
%                 \includegraphics[width=.65\linewidth, trim={1.1cm 1.1cm 0cm 0}, clip]{Act1_Iemocap.png}
%                 \caption{after DANN}
%                 \label{fig:gull2}
%         \end{subfigure}%
%         \caption{Illustration of data distributions from source and target before (a) and after (b) the DANN training. 
% Source:~\cite{Mohammed18-DAF}}\label{fig:animals}
% \end{figure}

%     \begin{figure}
% %         \begin{subfigure}[b]{0.50\textwidth}
%                 \centering
% %                 \includegraphics[width=.65\linewidth, trim={1.1cm 1.1cm 0cm 0}, clip]{Act1_Iemocap_src.png}
% %                 \caption{before DANN}
% %                 \label{fig:gull}
% %         \end{subfigure}%
% %         \vskip\baselineskip
%         \begin{subfigure}[b]{0.50\textwidth}
%                 \centering
%                 \includegraphics[width=.85\linewidth, trim={1.1cm 1.1cm 0cm 0}, clip]{Act1_Iemocap.png}
% %                 \caption{after DANN}
%                 \label{fig:gull2}
%         \end{subfigure}%
%         \caption{Illustration of data distributions from source and target after the DANN training. 
% Source:~\cite{Mohammed18-DAF}}\label{fig:animals}
% \end{figure}

\subsection{(Virtual) Adversarial Training} 
\label{subsec:virtualAdv}

\noindent
Beside factors relating to the quality of the training data, the performance of an emotional AI system  is also heavily dependent on its robustness to unseen data. A trivial disturbance on the sample (adversarial examples) might result in an opposite prediction~\cite{goodfellow2015explaining}, which naturally has to be avoided for a robust recognition model.

Generally speaking, \textit{adversarial examples} are the examples that are created by making small, but intentionally, perturbations to the input to incur large and significant perturbations in outputs (\eg incorrect predictions with high confidence)~\cite{goodfellow2015explaining}. \textit{Adversarial training}, however, addresses this vulnerability in recognition models by introducing mechanisms to correctly handle the adversarial examples. In this way, it improves not only robustness to adversarial examples, but also overall generalisation for the original examples~\cite{goodfellow2015explaining}.

Mathematically, adversarial training adds the following term as regularisation loss to the original loss function:
\begin{equation}
\label{eq:adversarialtraining1}
-\log p(y \vert {x}+{r}_{adv};\theta),
\end{equation}
in which ${x}$ denotes the input, $\theta$ denotes the parameters of a classifier, and ${r}_{adv}$ denotes a worst-case perturbation against the current model $p(y \vert {x};\theta)$, which can be calculated with 
\begin{equation}
\label{eq:adversarialtraining2}
{r}_{adv} = \arg\max_{{r}, \Vert {r} \Vert \leq \epsilon} \log p(y \vert {x}+{r};\theta).
\end{equation}

In the context of affective computing and sentiment analysis, the authors in~\cite{Chang17-LRO} utilised DCGAN and multi-task learning strategies to leverage a large number of unlabelled samples, where the unlabelled samples are considered as adversarial examples. More specifically, the model explores unlabelled data by feeding it through a vanilla DCGAN, in which a discriminator only learns to classify the input as either real or fake. Hence, no label information is demanded. Note that, the discriminator shares layers with another two classifiers to predict valence and activation simultaneously. This method has been shown to improve generalisability across corpora~\cite{Chang17-LRO}. A similar approach was conducted in~\cite{Deng17-SDO} to learn robust representations from emotional speech data for autism detection.

More recently, a cGAN-based framework was proposed for continuous speech emotion recognition in~\cite{Han18-TCA}, where a predictor and a discriminator are conditioned by acoustic features. In particular, the discriminator is employed to distinguish the joint probability distributions for acoustic features and their corresponding predictions or real annotations. In this way, the predictor is guided to modify the original predictions to achieve a better performance level.

Rather than the above mentioned adversarial training schemes that {\it explicitly} rely on the presence of a discriminator network, adversarial training can also be executed in an {\it implicit} manner, namely, {\it virtual adversarial training}. Virtual adversarial training is conducted by straightforwardly adding an additional regularisation term, which is sensitive to the adversarial examples, as a penalty in a loss function. 

In \textit{virtual adversarial training}, first proposed in~\cite{miyato2016distributional}, the regularisation loss term of  Equation~(\ref{eq:adversarialtraining1}) is reformulated without the label $y$ as follows: 
\begin{equation}
\label{eq:virtualadversarialtraining1}
\text{KL}[p(\cdot \vert {x};\theta)\Vert p(\cdot \vert {x}+{r}_{adv};\theta)],
\end{equation}
where $\text{KL}[\cdot \Vert \cdot]$ denotes the KL divergence between two distributions and a worst-case perturbation ${r}_{adv}$ can be computed by
\begin{equation}
\label{eq:virtualadversarialtraining2}
{r}_{adv} = \arg\max_{{r}, \Vert {r} \Vert \leq \epsilon}\text{KL}[p(\cdot \vert {x};\theta)\Vert p(\cdot \vert {x}+{r};\theta)].
\end{equation}

Inspired by these works, authors in~\cite{Miyato17-ATM} reported that, state-of-the-art sentiment classification results can be achieved when adopting (virtual) adversarial training approaches in the text domain. In particular, the authors applied perturbations to word embeddings in a recurrent neural network structure, rather than to the original input itself~\cite{Miyato17-ATM}. Following this success, (virtual) adversarial training has also been applied to speech emotion recognition in~\cite{Sahu18-SMP}. Results in~\cite{Sahu18-SMP} demonstrate that, the classification accuracy as well as the system's overall 
generalisation capability can be improved.

% {\blue
% Performance evaluation for emotion perception and understanding is more straightforward, comparing with other applications. Apart from visualising the 
% distribution of the generated data, the improvement achieved by the implementation of adversarial training can be reported on metrics such as unweighted 
% accuracy, unweighted average recall, and concordance correlation coefficient~\cite{Chen17-ADA, Deng17-SDO, Mohammed18-DAF}. Nevertheless, to quantify 
% the models' performance, a common evaluation metric designed specifically for this application is still missing for a fair comparison among various models.
% }

\begin{table*}[!t]
\caption{A summary of adversarial training studies in affective computing and sentiment analysis. These studies are listed by their applied tasks (SYN: 
synthesis, CVS: conversion, DA: data augmentation, DAT: domain adversarial training, AT: adversarial training, VAT: virtual adversarial training), modalities, 
and published years. GATH: generative adversarial talking head, ASPD: adversarial shared-private model. }
\centering
 \begin{threeparttable}
 \begin{tabular}{llllp{0.13\textwidth}p{0.39\textwidth}}
 \toprule
{\bf paper}                                        & {\bf year} & {\bf task} & {\bf modality}      & {\bf model}                   & {\bf note}                                                                                                                 \\
\midrule

Radfod~\et~\cite{Radford16-URL}                        & 2016 & SYN  & image         & DCGAN                   &  vector arithmetic can be done in 
latent vector space, \eg smiling woman - neutral woman + neutral man = smiling man \\
Chen~\et~\cite{Chen16-IIR}                             & 2016 & SYN  & image         & infoGAN                 & latent code can be interpreted; support gradual transformation                                                       \\
Huang\ \& Khan~\cite{Huang17-DGF}                        & 2017 & SYN  & image/video   & DyadGAN                 & interaction scenario; identity + attribute from the interviewee                                                        \\
Melis\ \& Amores~\cite{Melis17-TEG}                      & 2017 & SYN  & Image (art)   & cGAN                    & generate emotional artwork                                                                                           \\
Bao~\et~\cite{Bao18-TOI}                               & 2018 & SYN  & image         & identity preserving GAN & the identity and attributes of faces are separated                                                                   \\
Pham~\et~\cite{Pham18-GAT}                             & 2018 & SYN  & image/video   & GATH                    & conditioned by AU; from static image to dynamic video                                                                \\
Song~\et~\cite{Song18-TFG}                             & 2018 & SYN  & image/video   & cGAN                    & static image to dynamic video, conditioned by 
audio sequences                                                        \\
Nojavansghari~\et~\cite{Nojavansghari18-IGA}           & 2018 & SYN  & image         & DyadGAN (with RNN)      & interaction  scenario; smooth the video 
synthesis with context information                                             \\
Wang\ \& Arit{\`{e}}res~\cite{Wang18-MCS} & 2018 & SYN  & motion        & cGAN                    & to simulate a latent vector with seq2seq AE; 
controlled by emotion                                                   \\
Rajeswar~\et~\cite{Rajeswar17-AGO}                     & 2017 & SYN  & text          & (W)GAN-GP               & gradient penalty; generate sequences 
with fixed length                                                                \\
Liu~\et~\cite{Liu18-BND}                               & 2018 & SYN  & text (poetry) & I2P-GAN                 & multi-adversarial training; generate poetry 
from images                                                              \\
Fedus~\et~\cite{Fedus18-MBT}                           & 2018 & SYN  & text          & maskGAN                 & based on actor-critic cGAN; generate 
sequences 
with fixed length                                                     \\
\midrule
Perarnau~\et~\cite{Perarnau16-ICG}                     & 2016 & CVS  & image         & IcGAN                   & interpretable latent code; support gradual transformation                                                            \\
Zhou\ \& Shi~\cite{Zhou17-PFE}                           & 2017 & CVS  & image         & cGAN                    & learn the difference between the source and target emotions by adversarial autoencoder                               \\
Song~\et~\cite{Song17-GGA}                             & 2017 & CVS  & image         & G2-GAN                  &  geometry-guided, similar to cycleGAN    
                                                                              \\
Qiao~\et~\cite{Qiao18-GCG}                             & 2018 & CVS  & image         & GC-GAN                  & geometry-contrastive learning; the attribute 
and identity features are separated in the learning process             \\
Ding~\et~\cite{Ding18-EFE}                             & 2018 & CVS  & image         & exprGAN                 & intensity can be controlled                                                                                          \\
Lee~\et~\cite{Lee18-SSF}                               & 2018 & CVS  & text          & cycleGAN                & no need of paired data; emotion scalable                                                                             \\
Gao~\et~\cite{Gao18-VIU}                               & 2018 & CVS  & speech        & voiceGAN                & no need of paired data         \\
\midrule 
Zhu~\et~\cite{Zhu17-ECW}                               &2018 & DA   & image         & cycleGAN                & transfer data from A to B;  require no 
further labelling process on the transferred data                             \\
Sahu~\et~\cite{Sahu17-AAF}                             & 2017 & DA   & speech        & adversarial AE          & use GMM built on the original  data to label 
generated data -\textgreater noisy data; sensitive to mode collapse   \\
% \midrule
Ganin~\et~\cite{Ganin16-DAT}                           & 2016 & DAT  & text          & DANN                    & first work in domain  adversarial training                                                                           \\
Chen~\et~\cite{Chen17-ADA}                             & 2017 & DAT  & text          & DANN                    & semi-supervised supported;  Wasserstein distance used for smoothing training process                                 \\
Zhang~\et~\cite{Zhang17-AAN}                           & 2017 & DAT  & text          & DANN                    & attention scoring network is added for document embedding                                                            \\
Li~\et~\cite{Li17-EAM}                                 & 2017 & DAT  & text          & DANN                    & with attention mechanisms                      
                                                                      \\
Zhao~\et~\cite{Zhao18-MSD}                             & 2018 & DAT  & text          & multisource DANN        & extended for multiple sources           
                                                                              \\
Shen~\et~\cite{Shen18-WDG}                             & 2018 & DAT  & text          & DANN                    & Wasserstein distance guided to optimise domain discriminator                                                         \\
Liu~\et~\cite{Liu18-AML}                               & 2018 & DAT  & text          & ASPD                    & for multi-task; semi-supervised friendly                                                                             \\
Chen\ \& Cardie~\cite{Chen18-MAN}                        & 2018 & DAT  & text          & multinomial ASPD        & multinomial discriminator for 
multi-domain                                                                          \\
Mohammed\ \& Busso~\cite{Mohammed18-DAF}                 & 2018 & DAT  & speech        & DANN                    & adapted for speech emotion recognition        
                                                                     \\
% \midrule
Chang\ \& Scherer~\cite{Chang17-LRO}                     & 2017 & AT   & speech        & DCGAN                   & spectrograms with fixed width are randomly 
selected and chopped from a varied length of audio files                  \\
Deng~\et~\cite{Deng17-SDO}                             & 2017 & AT   & speech        & GAN                     & use hidden-layer representations from 
discriminator                                                               \\
Han~\et~\cite{Han18-TCA}                               & 2018 & AT   & speech        & cGAN                    & regularisation: joint distribution                                                                                   \\
Miyato~\et~\cite{Miyato17-ATM}                         & 2017 & VAT/AT  & text          & /                       & first work on virtual adversarial training                                                                           \\
Sahu~\et~\cite{Sahu18-SMP}                             & 2018 & VAT/AT  & speech        & DNN                     & first work for speech emotion recognition                                                                            \\% 
 \bottomrule
 \end{tabular}
 \end{threeparttable}
 \label{tab:summary}
\end{table*}

%%%%%%%%%%%%%%%%%%%%%%%%%%%%%%%%%%%%%%%%%%%%%%%%%%%%%%%%%%%%%%%%%%%%%%
\section{The Road Ahead}
\label{sec:futureWork}

% While much progress has been made in adversarial training to alleviate some of the challenges related affective computing and sentiment analysis, for example, synthesising and transforming image-based emotion through cycleGAN, augmenting data by artificially generating samples, extracting of robust representations via domain adversarial training, it still requires several breakthroughs in both the theorem of adversarial training and its application to the affective computing and sentiment analysis. %\nc{Draw attention here to the promising approaches already mentioned in the previous sections}

\noindent
Considerable progress has been made in alleviating some of the challenges related to affective computing and sentiment analysis through the use of adversarial training, for example, synthesising and transforming image-based emotions through cycleGAN, augmenting data by artificially generating samples, extracting robust representations via domain adversarial training. A detailed summary of these works can be found in Table~\ref{tab:summary}. However, large scale breakthroughs are still required in both the theorem of adversarial training and its applications to fully realise the potential of this paradigm in affective computing and sentiment analysis. 

\subsection{Limitations of Adversarial Training}
% The arguably two major problems of adversarial training come to the training instability and the model collapse, which remain open-ended research directions in adversarial training. Solving these fundamental problems will certainty facilitate its application to affective computing and sentiment analysis. 
% \nc{again, include a line or two to link back to affective/sentiment - why is it important for readers from these fields to know the below points}
\noindent
Arguably, the two major open research challenges relating to adversarial training are \emph{training instability} and \emph{mode collapse}. Solving these fundamental concerns will help facilitate its application to affective computing and sentiment analysis.
% \nc{Solving these will help in application that uses adversarial training, not just affect and sentiment, is it worth saying this?}. 

\subsubsection{Training Instability} 
In the adversarial training process, ensuring that there is balance and synchronization between the two adversarial networks plays an important role in obtaining reliable results~\cite{Goodfellow14-GAN}. That is, the goal optimisation of adversarial training lies in finding a saddle point of, rather than a local minimum between, the two adversarial components. The inherent difficulty in controlling the synchronisation of the two adversarial networks increases the risk of instability in the training process.

To date, researchers have made several attempts to address this problem. For example, the implementation of Wasserstein distance rather than the conventional JS divergence partially solves the vanishing gradient problem associated with improvements in the ability of the discriminator to separate the real and generated samples~\cite{Arjovsky17-WG}. Furthermore, the convergence of the model and the existence of the equilibrium point have yet to be theoretically proven~\cite{Arora17-GAE}. Therefore, it remains an open research direction to further optimise the training process. 
% \ncedit{\st{it} issues relating to balance and synchronization} remain\ncedit{\st{s}} an open research direction to further optimise the training process. 

\subsubsection{Mode Collapse} 
% Model collapse indicates that the generated samples smash into a small family of similar samples (partial collapse), or even a single sample (complete collapse). In this case, the generator will exhibit very limited diversity amongst generated samples, which reduces the usefulness of the learnt GANs. 

% Novel approaches dedicated to solve the model collapse problem  are continually emerging. For example, the cost function of the generator can be modified to factor in the diversity of generated batches~\cite{Salimans16-ITF}, the unroll GAN allows the generator to ``unroll'' updates of the discriminator in a fully differentiable way~\cite{Metz16-UGA}, and the AdaGAN combines an ensemble of GANs~\cite{Tolstikhin17-ABG}. 

Mode collapse occurs when the the generator exhibits very limited diversity among generated samples, thus reducing the usefulness of the learnt GANs. This effect can be observed as the generated samples can be integrated into a small subset of similar samples (partial collapse), or even a single sample (complete collapse).

Novel approaches dedicated to solving the mode collapse problem are continually emerging. For example, the loss function of the generator can be  modified to factor in the diversity of generated samples in batches~\cite{Salimans16-ITF}. Alternatively, the unroll-GAN allows the generator to `unroll' the updates of the discriminator in a manner which is fully differentiable~\cite{Metz16-UGA}, and the AdaGAN combines an ensemble of GANs in a boosting framework to ensure diversity~\cite{Tolstikhin17-ABG}.

\subsection{Other Ongoing Breakthroughs}

\noindent
In most conditional GAN frameworks, the emotional entity is generated by utilising discrete emotional categories as the condition controller. However, emotions are more than these basic categories (\eg Ekman's Six Basic Emotions), and to date, more subtle emotional expressions have been largely overlooked in the literature. While some studies have started addressing this issue in image processing studies (cf.~Section~\ref{sec:synthesis}), it is still one of the major research white spots in speech and text processing. Therefore, using a more soft condition to replace the controller remains an open research direction. 

To the best of our knowledge, GAN-based emotional speech synthesis has yet to be addressed in the relevant literature. This could be due in part to Speech Emotion Recognition (SER) being a less mature field of research, which leads to a limited capability to distinguish the emotions using speech and thus provides deductible contributions to optimise the generator. However, with the ongoing development of SER and the already discussed success of GANs in conventional speech synthesis~\cite{Yang17-SPS} and image/video and text generation (cf.~Section~\ref{sec:synthesis}), we strongly believe that major breakthroughs will be made in this area sometime in the near future. 

Similarly, current state-of-the-art emotion conversion systems are based on the transformation of static images. However, transforming emotions in the dynamic sequential signals, such as speech, video, and text, remains challenging. This most likely relates to the difficulties associated with sequence-based discriminator and sequence generation. However, the state-of-the-art performance achieved with generative models, such as WaveGAN, indicate that adversarial training can play a key role in helping {to break} through these barriers.

Additionally, when comparing the performance of different GAN-based models, a fair comparison is vital but not straightforward. In~\cite{Liu18-BND}, the authors demonstrated that their proposed I2P-GAN outperforms SeqGAN when generating poetry from given images, reporting higher scores on evaluation metrics including BLEU, novelty, and relevance. Also, it has been claimed that InfoGAN converges faster than a conventional GAN framework~\cite{Chen16-IIR}. However, it should be noted that, a fair experimental comparison of various generative adversarial training models associated with affective computing and sentiment analysis has yet to be reported, to answer questions such as which model is faster, more accurate, or easier to implement. This absence is mainly due to the lack of benchmark datasets and thoughtfully designed metrics for each specific application (\ie emotion generation, emotion conversion, and emotion perception and understanding).

Finally, we envisage that, GAN-based end-to-end emotional dialogue systems can succeed the speech-to-text (\ie ASR) and the text-to-speech (\ie TTS) processes currently used in conventional dialogue systems. This is motivated by the  construct that humans generally do not consciously convert speech into text during conversations~\cite{Dunbar97-HCB}. The advantage of this approach is that it avoids the risk of the possible information loss during this internal 
process. Such an end-to-end emotional framework would further facilitate the next generation of more human-like dialogue systems. 

%%%%%%%%%%%%%%%%%%%%%%%%%%%%%%%%%%%%%%%%%%%%%%%%%%%%%%%%%%%%%%%%%%%%%%
\section{Conclusion}
\label{sec:conclusion}
% Motivated by the marvellous achievement made by adversarial training in artificial intelligence, in this article we systematically summarise the most recent advances of adversarial training in affective computing and sentiment analysis, from the emotion analysis and understanding to the emotion synthesis and conversion, and from audio, over image/video, to text modalities. Generally speaking, not only have the adversarial training made great contributions to the development of corresponding generative models, but they are also helpful and instructive for related discriminative models.

% We also have to be noted that further efforts as discussed in Section~\ref{sec:futureWork} have to be made before affective computing and sentiment analysis can be considered ready for broad consumer usage ``in the wild''. 
% With the development of adversarial training research, we believe that more provocative research of adversarial training will be made for affective computing and sentiment analysis in the near future.

\noindent
Motivated by the ongoing success and achievements associated with adversarial training in artificial intelligence, this article summarised the most recent advances of adversarial training in affective computing and sentiment analysis. Covering the audio, image/video, and text modalities, this overview included technologies and paradigms relating to both emotion synthesis and conversion as well as emotion perception and understanding. Generally speaking, not only have adversarial training techniques made great contributions to the development of corresponding generative models, but they are also helpful and instructive for related discriminative models. We have also drawn attention to further research efforts aimed at leveraging the highlighted advantages of adversarial training. If successfully implemented, such techniques will inspire and foster the new generation of robust affective computing and sentiment analysis technologies that are capable of widespread in-the-wild deployment.

\section*{Acknowledgment}
\noindent
This work has been supported by the EU's Horizon 2020 Programme through the Innovation Action No.~645094 (SEWA), the EU's Horizon 2020 / EFPIA Innovative 
Medicines Initiative through GA No.~115902 (RADAR-CNS), and the UK's Economic \& Social Research Council through the research Grant No.~HJ-253479 (ACLEW).

\ifCLASSOPTIONcaptionsoff
  \newpage
\fi

\balance
\bibliographystyle{IEEEtran}
\bibliography{cim_strong,cim_weak}

\begin{thebibliography}{100}
\providecommand{\url}[1]{#1}
\csname url@samestyle\endcsname
\providecommand{\newblock}{\relax}
\providecommand{\bibinfo}[2]{#2}
\providecommand{\BIBentrySTDinterwordspacing}{\spaceskip=0pt\relax}
\providecommand{\BIBentryALTinterwordstretchfactor}{4}
\providecommand{\BIBentryALTinterwordspacing}{\spaceskip=\fontdimen2\font plus
\BIBentryALTinterwordstretchfactor\fontdimen3\font minus
  \fontdimen4\font\relax}
\providecommand{\BIBforeignlanguage}[2]{{%
\expandafter\ifx\csname l@#1\endcsname\relax
\typeout{** WARNING: IEEEtran.bst: No hyphenation pattern has been}%
\typeout{** loaded for the language `#1'. Using the pattern for}%
\typeout{** the default language instead.}%
\else
\language=\csname l@#1\endcsname
\fi
#2}}
\providecommand{\BIBdecl}{\relax}
\BIBdecl

\bibitem{Picard97-AC}
R.~Picard, \emph{Affective computing}.\hskip 1em plus 0.5em minus 0.4em\relax
  Cambridge, MA: MIT Press, 1997.

\bibitem{Minsky07-TEM}
M.~Minsky, \emph{The emotion machine: {Commonsense} thinking, artificial
  intelligence, and the future of the human mind}.\hskip 1em plus 0.5em minus
  0.4em\relax New York, NY: Simon and Schuster, 2007.

\bibitem{Pantic05-AMH}
M.~Pantic, N.~Sebe, J.~F. Cohn, and T.~Huang, ``Affective multimodal
  human--computer interaction,'' in \emph{Proc.\ 13th ACM International
  Conference on Multimedia (MM)}, Singapore, 2005, pp. 669--676.

\bibitem{Poria15-SDF}
S.~Poria, E.~Cambria, A.~Gelbukh, F.~Bisio, and A.~Hussain, ``Sentiment data
  flow analysis by means of dynamic linguistic patterns,'' \emph{IEEE
  Computational Intelligence Magazine}, vol.~10, no.~4, pp. 26--36, Nov. 2015.

\bibitem{Poria15-TAI}
S.~Poria, E.~Cambria, A.~Hussain, and G.-B. Huang, ``Towards an intelligent
  framework for multimodal affective data analysis,'' \emph{Neural Networks},
  vol.~63, pp. 104--116, Mar. 2015.

\bibitem{Cambria16-ACA}
E.~Cambria, ``Affective computing and sentiment analysis,'' \emph{IEEE
  Intelligent Systems}, vol.~31, no.~2, pp. 102--107, Mar. 2016.

\bibitem{Chen16-LUA}
T.~Chen, R.~Xu, Y.~He, Y.~Xia, and X.~Wang, ``Learning user and product
  distributed representations using a sequence model for sentiment analysis,''
  \emph{IEEE Computational Intelligence Magazine}, vol.~11, no.~3, pp. 34--44,
  Aug. 2016.

\bibitem{Han17-SMF}
J.~Han, Z.~Zhang, N.~Cummins, F.~Ringeval, and B.~Schuller, ``Strength
  modelling for real-world automatic continuous affect recognition from
  audiovisual signals,'' \emph{Image and Vision Computing}, vol.~65, pp.
  76--86, Sep. 2017.

\bibitem{Santos14-DCN}
C.~N. dos Santos and M.~Gatti, ``Deep convolutional neural networks for
  sentiment analysis of short texts,'' in \emph{Proc.\ 25th International
  Conference on Computational Linguistics (COLING)}, Dublin, Ireland, 2014, pp.
  69--78.

\bibitem{Tzirakis17-EME}
P.~Tzirakis, G.~Trigeorgis, M.~A. Nicolaou, B.~Schuller, and S.~Zafeiriou,
  ``End-to-end multimodal emotion recognition using deep neural networks,''
  \emph{IEEE Journal of Selected Topics in Signal Processing, Special Issue on
  End-to-End Speech and Language Processing}, vol.~11, no.~8, pp. 1301--1309,
  Dec. 2017.

\bibitem{Zhang17-ADE}
Z.~Zhang, N.~Cummins, and B.~Schuller, ``Advanced data exploitation for speech
  analysis -- an overview,'' \emph{IEEE Signal Processing Magazine}, vol.~34,
  no.~4, pp. 107--129, July 2017.

\bibitem{Chen17-ADA}
X.~Chen, Y.~Sun, B.~Athiwaratkun, C.~Cardie, and K.~Weinberger, ``Adversarial
  deep averaging networks for cross-lingual sentiment classification,''
  \emph{arXiv preprint arXiv:1606.01614}, Apr. 2017.

\bibitem{Deng18-SAF}
J.~Deng, X.~Xu, Z.~Zhang, S.~Fr{\"u}hholz, and B.~Schuller, ``Semisupervised
  autoencoders for speech emotion recognition,'' \emph{IEEE/ACM Transactions on
  Audio, Speech, and Language Processing}, vol.~26, no.~1, pp. 31--43, Jan.
  2018.

\bibitem{Rajeswar17-AGO}
S.~Subramanian, S.~Rajeswar, F.~Dutil, C.~Pal, and A.~C. Courville,
  ``Adversarial generation of natural language,'' in \emph{Proc.\ 2nd Workshop
  on Representation Learning for NLP (Rep4NLP@ACL)}, Vancouver, Canada, 2017,
  pp. 241--251.

\bibitem{Gao18-VIU}
Y.~Gao, R.~Singh, and B.~Raj, ``Voice impersonation using generative
  adversarial networks,'' in \emph{Proc.\ IEEE International Conference on
  Acoustics, Speech and Signal Processing (ICASSP)}, Calgary, Canada, 2018, pp.
  2506--2510.

\bibitem{Goodfellow14-GAN}
I.~Goodfellow, J.~Pouget-Abadie, M.~Mirza, B.~Xu, D.~Warde-Farley, S.~Ozair,
  A.~Courville, and Y.~Bengio, ``Generative adversarial nets,'' in \emph{Proc.\
  Advances in Neural Information Processing Systems (NIPS)}, Montreal, Canada,
  2014, pp. 2672--2680.

\bibitem{Wang17-GAN}
K.~Wang, C.~Gou, Y.~Duan, Y.~Lin, X.~Zheng, and F.-Y. Wang, ``Generative
  adversarial networks: {Introduction} and outlook,'' \emph{IEEE/CAA Journal of
  Automatica Sinica}, vol.~4, no.~4, pp. 588--598, Sep. 2017.

\bibitem{Creswell18-GAN}
A.~Creswell, T.~White, V.~Dumoulin, K.~Arulkumaran, B.~Sengupta, and A.~A.
  Bharath, ``Generative adversarial networks: An overview,'' \emph{IEEE Signal
  Processing Magazine}, vol.~35, no.~1, pp. 53--65, Jan. 2018.

\bibitem{Radford16-URL}
A.~Radford, L.~Metz, and S.~Chintala, ``Unsupervised representation learning
  with deep convolutional generative adversarial networks,'' San Juan, PR,
  2016.

\bibitem{Han18-TCA}
J.~Han, Z.~Zhang, Z.~Ren, F.~Ringeval, and B.~Schuller, ``Towards conditional
  adversarial training for predicting emotions from speech,'' in \emph{Proc.\
  IEEE International Conference on Acoustics, Speech and Signal Processing
  (ICASSP)}, Calgary, Canada, 2018, pp. 6822--6826.

\bibitem{Fedus18-MBT}
W.~Fedus, I.~Goodfellow, and A.~M. Dai, ``{MaskGAN: B}etter text generation via
  filling in the \_,'' in \emph{Proc. 6th International Conference on Learning
  Representations (ICLR)}, Vancouver, Canada, 2018.

\bibitem{Zeng09-ASO}
Z.~Zeng, M.~Pantic, G.~I. Roisman, and T.~S. Huang, ``A survey of affect
  recognition methods: {Audio}, visual, and spontaneous expressions,''
  \emph{IEEE Transactions on Pattern Analysis and Machine Intelligence},
  vol.~31, no.~1, pp. 39--58, Jan. 2009.

\bibitem{Calvo10-ADA}
R.~A. Calvo and S.~D'Mello, ``Affect detection: {An} interdisciplinary review
  of models, methods, and their applications,'' \emph{IEEE Transactions on
  Affective Computing}, vol.~1, no.~1, pp. 18--37, Jan. 2010.

\bibitem{Gunes11-ERA}
H.~Gunes, B.~Schuller, M.~Pantic, and R.~Cowie, ``Emotion representation,
  analysis and synthesis in continuous space: {A survey},'' in \emph{Proc.\
  IEEE International Conference on Automatic Face \& Gesture Recognition and
  Workshops (FG)}, Santa Barbara, CA, 2011, pp. 827--834.

\bibitem{Schuller18-SER}
B.~Schuller, ``Speech emotion recognition: {Two} decades in a nutshell,
  benchmarks, and ongoing trends,'' \emph{Communications of the ACM}, vol.~61,
  no.~5, pp. 90--99, Apr. 2018.

\bibitem{Liu12-SAA}
B.~Liu, \emph{Sentiment analysis and opinion mining}.\hskip 1em plus 0.5em
  minus 0.4em\relax San Rafael, CA: Morgan \& Claypool Publishers, 2012.

\bibitem{Medhat14-SAA}
W.~Medhat, A.~Hassan, and H.~Korashy, ``Sentiment analysis algorithms and
  applications: {A} survey,'' \emph{Ain Shams Engineering Journal}, vol.~5,
  no.~4, pp. 1093--1113, Dec. 2014.

\bibitem{Liu15-SAM}
B.~Liu, \emph{Sentiment analysis: {Mining} opinions, sentiments, and
  emotions}.\hskip 1em plus 0.5em minus 0.4em\relax Cambridge, United Kingdom:
  Cambridge University Press, June 2015.

\bibitem{Cambria17-SAI}
E.~Cambria, S.~Poria, A.~Gelbukh, and M.~Thelwall, ``Sentiment analysis is a
  big suitcase,'' \emph{IEEE Intelligent Systems}, vol.~32, no.~6, pp. 74--80,
  Dec. 2017.

\bibitem{Poria17-ARO}
S.~Poria, E.~Cambria, R.~Bajpai, and A.~Hussain, ``A review of affective
  computing: {From} unimodal analysis to multimodal fusion,'' \emph{Information
  Fusion}, vol.~37, pp. 98--125, Sep. 2017.

\bibitem{Soleymani17-ASO}
M.~Soleymani, D.~Garcia, B.~Jou, B.~Schuller, S.-F. Chang, and M.~Pantic, ``{A
  Survey of Multimodal Sentiment Analysis},'' \emph{Image and Vision
  Computing}, vol.~65, pp. 3--14, Sep. 2017.

\bibitem{Zhang18-DLFS}
L.~Zhang, S.~Wang, and B.~Liu, ``Deep learning for sentiment analysis: {A}
  survey,'' \emph{Wiley Interdisciplinary Reviews: Data Mining and Knowledge
  Discovery}, Mar. 2018, 25 pages.

\bibitem{Oord16-WAG}
A.~van~den Oord, S.~Dieleman, H.~Zen, K.~Simonyan, O.~Vinyals, A.~Graves,
  N.~Kalchbrenner, A.~W. Senior, and K.~Kavukcuoglu, ``{WaveNet}: {A}
  generative model for raw audio,'' \emph{arXiv preprint arXiv:1609.03499},
  Sep. 2016.

\bibitem{Oord16-Oord}
A.~van~den Oord, N.~Kalchbrenner, and K.~Kavukcuoglu, ``Pixel recurrent neural
  networks,'' in \emph{Proc.\ 33rd International Conference on Machine Learning
  (ICML)}, New York City, NY, 2016, pp. 1747--1756.

\bibitem{Kingma13-AEV}
D.~P. Kingma and M.~Welling, ``Auto-encoding variational {Bayes},'' \emph{arXiv
  preprint arXiv:1312.6114}, Dec. 2013.

\bibitem{Lee17-EET}
Y.~Lee, A.~Rabiee, and S.-Y. Lee, ``Emotional end-to-end neural speech
  synthesizer,'' \emph{arXiv preprint arXiv:1711.05447}, Nov. 2017.

\bibitem{Akuzawa18-ESS}
K.~Akuzawa, Y.~Iwasawa, and Y.~Matsuo, ``Expressive speech synthesis via
  modeling expressions with variational autoencoder,'' in \emph{Proc.\ Annual
  Conference of the International Speech Communication Association
  (INTERSPEECH)}, Hyderabad, India.

\bibitem{Qiao18-GCG}
F.~Qiao, N.~Yao, Z.~Jiao, Z.~Li, H.~Chen, and H.~Wang, ``Geometry-contrastive
  generative adversarial network for facial expression synthesis,'' \emph{arXiv
  preprint arXiv:1802.01822}, Feb. 2018.

\bibitem{Zhou17-ECM}
H.~Zhou, M.~Huang, T.~Zhang, X.~Zhu, and B.~Liu, ``Emotional chatting machine:
  {Emotional} conversation generation with internal and external memory,'' in
  \emph{Proc.\ 32nd Conference on Association for the Advancement of Artificial
  Intelligence (AAAI)}, New Orleans, LA, 2018, pp. 730--738.

\bibitem{Schuller10-CCA}
B.~Schuller, B.~Vlasenko, F.~Eyben, M.~W\"ollmer, A.~Stuhlsatz, A.~Wendemuth,
  and G.~Rigoll, ``Cross-corpus acoustic emotion recognition: Variances and
  strategies,'' \emph{IEEE Transactions on Affective Computing}, vol.~1, no.~2,
  pp. 119--131, July 2010.

\bibitem{Han17-FHT}
J.~Han, Z.~Zhang, M.~Schmitt, M.~Pantic, and B.~Schuller, ``From hard to soft:
  {Towards} more human-like emotion recognition by modelling the perception
  uncertainty,'' in \emph{Proc.\ 25th ACM International Conference on
  Multimedia (MM)}, Mountain View, CA, 2017, pp. 890--897.

\bibitem{Kim13-DLF}
Y.~Kim, H.~Lee, and E.~M. Provost, ``Deep learning for robust feature
  generation in audiovisual emotion recognition,'' in \emph{Proc.\ IEEE
  International Conference on Acoustics, Speech and Signal Processing
  (ICASSP)}, Vancouver, Canada, 2013, pp. 3687--3691.

\bibitem{Cummins17-AID}
N.~Cummins, S.~Amiriparian, G.~Hagerer, A.~Batliner, S.~Steidl, and
  B.~Schuller, ``An image-based deep spectrum feature representation for the
  recognition of emotional speech,'' in \emph{Proc.\ 25th ACM International
  Conference on Multimedia (MM)}, Mountain View, CA, 2017, pp. 478--484.

\bibitem{Zhang18-LUD}
Z.~Zhang, J.~Han, J.~Deng, X.~Xu, F.~Ringeval, and B.~Schuller, ``Leveraging
  unlabelled data for emotion recognition with enhanced collaborative
  semi-supervised learning,'' \emph{IEEE Access}, vol.~6, pp. 22\,196--22\,209,
  Apr. 2018.

\bibitem{Pan10-ASO}
S.~J. Pan and Q.~Yang, ``A survey on transfer learning,'' \emph{IEEE
  Transactions on Knowledge and Data Engineering}, vol.~22, no.~10, pp.
  1345--1359, Oct. 2010.

\bibitem{Glorot11-DAF}
X.~Glorot, A.~Bordes, and Y.~Bengio, ``Domain adaptation for large-scale
  sentiment classification: {A} deep learning approach,'' in \emph{Proc.\ 28th
  International Conference on Machine Learning (ICML)}, Bellevue, WA, 2011, pp.
  513--520.

\bibitem{Deng14-ISA}
J.~Deng, R.~Xia, Z.~Zhang, Y.~Liu, and B.~Schuller, ``Introducing
  shared-hidden-layer autoencoders for transfer learning and their application
  in acoustic emotion recognition,'' in \emph{Proc.\ IEEE International
  Conference on Acoustics, Speech and Signal Processing (ICASSP)}, Florence,
  Italy, 2014, pp. 4818--4822.

\bibitem{You15-RIS}
Q.~You, J.~Luo, H.~Jin, and J.~Yang, ``Robust image sentiment analysis using
  progressively trained and domain transferred deep networks,'' in \emph{Proc.\
  29th Conference on Association for the Advancement of Artificial Intelligence
  (AAAI)}, Austin, TX, 2015, pp. 381--388.

\bibitem{Bao18-TOI}
J.~Bao, D.~Chen, F.~Wen, H.~Li, and G.~Hua, ``Towards open-set identity
  preserving face synthesis,'' in \emph{Proc.\ {IEEE} Conference on Computer
  Vision and Pattern Recognition (CVPR)}, Salt Lake City, UT, 2018, pp.
  6713--6722.

\bibitem{Nojavansghari18-IGA}
B.~Nojavanasghari, Y.~Huang, and S.~Khan, ``Interactive generative adversarial
  networks for facial expression generation in dyadic interactions,''
  \emph{arXiv preprint arXiv:1801.09092}, Jan. 2018.

\bibitem{Huang17-DGF}
Y.~Huang and S.~M. Khan, ``{DyadGAN: Generating} facial expressions in dyadic
  interactions,'' in \emph{Proc.\ {IEEE} Conference on Computer Vision and
  Pattern Recognition Workshops (CVPRW)}, Honolulu, HI, 2017, pp. 2259--2266.

\bibitem{Papineni02-BLEU}
K.~Papineni, S.~Roukos, T.~Ward, and W.-J. Zhu, ``{BLEU: A} method for
  automatic evaluation of machine translation,'' in \emph{Proc.\ 40th Annual
  Meeting of the Association for Computational Linguistics (ACL)}, Stroudsburg,
  PA, 2002, pp. 311--318.

\bibitem{lin2004rouge}
C.-Y. LIN, ``{ROUGE: A }package for automatic evaluation of summaries,'' in
  \emph{Proc. Text Summarization Branches Out Workshop in ACL}, Barcelona,
  Spain, 2004, 8 pages.

\bibitem{Deng17-SDO}
J.~Deng, N.~Cummins, M.~Schmitt, K.~Qian, F.~Ringeval, and B.~Schuller,
  ``Speech-based diagnosis of autism spectrum condition by generative
  adversarial network representations,'' in \emph{Proc.\ International
  Conference on Digital Health (DH)}, London, UK, 2017, pp. 53--57.

\bibitem{Mohammed18-DAF}
M.~Abdelwahab and C.~Busso, ``Domain adversarial for acoustic emotion
  recognition,'' \emph{IEEE/ACM Transactions on Audio, Speech, and Language
  Processing}, vol.~26, no.~12, pp. 2423--2435, Dec. 2018.

\bibitem{Caesar17-gan}
H.~Caesar, ``Really-awesome-gan,''
  \url{https://github.com/nightrome/really-awesome-gan}, 2017.

\bibitem{Hindupur18-gan}
A.~Hindupur, ``The-gan-zoo,''
  \url{https://github.com/hindupuravinash/the-gan-zoo}, 2018.

\bibitem{Arjovsky17-WG}
M.~Arjovsky, S.~Chintala, and L.~Bottou, ``Wasserstein {GAN},'' \emph{arXiv
  preprint arXiv:1701.07875}, Mar. 2017.

\bibitem{zhao2016energy}
J.~Zhao, M.~Mathieu, and Y.~LeCun, ``Energy-based generative adversarial
  network,'' in \emph{Proc.\ 5th International Conference on Learning
  Representations (ICLR)}, Toulon, France, 2017.

\bibitem{mao2017least}
X.~Mao, Q.~Li, H.~Xie, R.~Y. Lau, Z.~Wang, and S.~Paul~Smolley, ``Least squares
  generative adversarial networks,'' in \emph{Proc.\ IEEE International
  Conference on Computer Vision (ICCV)}, Venice, Italy, 2017, pp. 2794--2802.

\bibitem{qi2017loss}
G.-J. Qi, ``Loss-sensitive generative adversarial networks on {Lipschitz}
  densities,'' \emph{arXiv preprint arXiv:1701.06264}, Jan. 2017.

\bibitem{patel2018correlated}
S.~Patel, A.~Kakadiya, M.~Mehta, R.~Derasari, R.~Patel, and R.~Gandhi,
  ``Correlated discrete data generation using adversarial training,''
  \emph{arXiv preprint arXiv:1804.00925}, 2018.

\bibitem{che2016mode}
T.~Che, Y.~Li, A.~P. Jacob, Y.~Bengio, and W.~Li, ``Mode regularized generative
  adversarial networks,'' in \emph{Proc.\ 5th International Conference on
  Learning Representations (ICLR)}, Toulon, France, 2017.

\bibitem{Mirza14-CGA}
M.~Mirza and S.~Osindero, ``Conditional generative adversarial nets,''
  \emph{arXiv preprint arXiv:1411.1784}, June 2014.

\bibitem{odena2016semi}
A.~Odena, ``Semi-supervised learning with generative adversarial networks,'' in
  \emph{Proc.\ Data-Efficient Machine Learning Workshop in ICML}, New York, NY,
  2016.

\bibitem{donahue2016adversarial}
J.~Donahue, P.~Kr{\"a}henb{\"u}hl, and T.~Darrell, ``Adversarial feature
  learning,'' in \emph{Proc.\ 5th International Conference on Learning
  Representations (ICLR)}, Toulon, France, 2017.

\bibitem{Zhu17-UIT}
J.-Y. Zhu, T.~Park, P.~Isola, and A.~A. Efros, ``Unpaired image-to-image
  translation using cycle-consistent adversarial networks,'' in \emph{Proc.\
  IEEE Conference on Computer Vision and Pattern Recognition (ICCV)}, Venice,
  Italy, 2017, pp. 2223--2232.

\bibitem{pmlr-v70-kim17a}
T.~Kim, M.~Cha, H.~Kim, J.~K. Lee, and J.~Kim, ``Learning to discover
  cross-domain relations with generative adversarial networks,'' in
  \emph{Proc.\ 34th International Conference on Machine Learning (ICML)},
  Sydney, Australia, 2017, pp. 1857--1865.

\bibitem{Chen16-IIR}
X.~Chen, Y.~Duan, R.~Houthooft, J.~Schulman, I.~Sutskever, and P.~Abbeel,
  ``{InfoGAN: I}nterpretable representation learning by information maximizing
  generative adversarial nets,'' in \emph{Proc.\ Advances in Neural Information
  Processing Systems (NIPS)}, Barcelona, Spain, 2016, pp. 2172--2180.

\bibitem{chongxuan2017triple}
L.~Chongxuan, T.~Xu, J.~Zhu, and B.~Zhang, ``Triple generative adversarial
  nets,'' in \emph{Proc.\ Advances in Neural Information Processing Systems
  (NIPS)}, Long Beach, CA, 2017, pp. 4091--4101.

\bibitem{luo2017learning}
J.~Luo, Y.~Xu, C.~Tang, and J.~Lv, ``Learning inverse mapping by autoencoder
  based generative adversarial nets,'' in \emph{Proc.\ International Conference
  on Neural Information Processing (ICONIP)}, Guangzhou, China, 2017, pp.
  207--216.

\bibitem{mogren2016c}
O.~Mogren, ``{C-RNN-GAN: Continuous} recurrent neural networks with adversarial
  training,'' in \emph{Proc.\ Constructive Machine Learning Workshop in NIPS},
  Barcelona, Spain, 2016, 6 pages.

\bibitem{xu2017attngan}
T.~Xu, P.~Zhang, Q.~Huang, H.~Zhang, Z.~Gan, X.~Huang, and X.~He, ``{AttnGAN:
  Fine}-grained text to image generation with attentional generative
  adversarial networks,'' in \emph{Proc.\ {IEEE} Conference on Computer Vision
  and Pattern Recognition (CVPR)}, Salt Lake City, UT, 2018, pp. 1316--1324.

\bibitem{jaiswal2018capsulegan}
A.~Jaiswal, W.~AbdAlmageed, and P.~Natarajan, ``{CapsuleGAN: Generative}
  adversarial capsule network,'' \emph{arXiv preprint arXiv:1802.06167}, Feb.
  2018.

\bibitem{creswell2016adversarial}
A.~Creswell and A.~A. Bharath, ``Adversarial training for sketch retrieval,''
  in \emph{Proc.\ European Conference on Computer Vision (ECCV)}, Amsterdam,
  Netherlands, 2016, pp. 798--809.

\bibitem{Tan17-AAS}
W.~R. Tan, C.~S. Chan, H.~E. Aguirre, and K.~Tanaka, ``{ArtGAN: Artwork}
  synthesis with conditional categorical {GANs},'' in \emph{Proc.\ IEEE
  International Conference on Image Processing (ICIP)}, Beijing, China, 2017,
  pp. 3760--3764.

\bibitem{pascual2017segan}
S.~Pascual, A.~Bonafonte, and J.~Serr{\`a}, ``{SEGAN: Speech} enhancement
  generative adversarial network,'' in \emph{Proc.\ 18th Annual Conference of
  the International Speech Communication Association (INTERSPEECH)}, Stockholm,
  Sweden, 2017, pp. 3642--3646.

\bibitem{donahue2018synthesizing}
C.~Donahue, J.~McAuley, and M.~Puckette, ``Synthesizing audio with generative
  adversarial networks,'' in \emph{Proc.\ Workshop in 6th International
  Conference on Learning Representations (ICLR)}, Vancouver, Canada, 2018.

\bibitem{Oord16-CIG}
A.~van~den Oord, N.~Kalchbrenner, L.~Espeholt, O.~Vinyals, A.~Graves, and
  K.~Kavukcuoglu, ``Conditional image generation with {PixelCNN} decoders,'' in
  \emph{Proc.\ Advances in Neural Information Processing Systems (NIPS)},
  Barcelona, Spain, 2016, pp. 4790--4798.

\bibitem{Caswell15-EAL}
I.~Caswell, O.~Sen, and A.~Nie, ``Exploring adversarial learning on neural
  network models for text classification,'' 2015.

\bibitem{Pham18-GAT}
H.~X. Pham, Y.~Wang, and V.~Pavlovic, ``Generative adversarial talking head:
  {Bringing} portraits to life with a weakly supervised neural network,''
  \emph{arXiv preprint arXiv:1803.07716}, Mar. 2018.

\bibitem{Song18-TFG}
Y.~Song, J.~Zhu, X.~Wang, and H.~Qi, ``Talking face generation by conditional
  recurrent adversarial network,'' \emph{arXiv preprint arXiv:1804.04786}, Apr.
  2018.

\bibitem{Melis17-TEG}
D.~Alvarez-Melis and J.~Amores, ``The emotional {GAN: Priming} adversarial
  generation of art with emotion,'' in \emph{Proc.\ Advances in Neural
  Information Processing Systems (NIPS)}, Long Beach, CA, 2017, 4 pages.

\bibitem{Wang18-MCS}
Q.~Wang and T.~Arti{\`{e}}res, ``Motion capture synthesis with adversarial
  learning,'' in \emph{Proc.\ 17th International Conference on Intelligent
  Virtual Agents (IVA)}, Stockholm, Sweden, 2017, pp. 467--470.

\bibitem{Liu18-BND}
B.~Liu, J.~Fu, M.~P. Kato, and M.~Yoshikawa, ``Beyond narrative description:
  {Generating} poetry from images by multi-adversarial training,'' \emph{arXiv
  preprint arXiv:1804.08473}, Apr. 2018.

\bibitem{Yu17-SSG}
L.~Yu, W.~Zhang, J.~Wang, and Y.~Yu, ``{SeqGAN}: {Sequence} generative
  adversarial nets with policy gradient.'' in \emph{Proc.\ 31st Conference on
  Association for the Advancement of Artificial Intelligence (AAAI)}, San
  Francisco, CA, 2017, pp. 2852--2858.

\bibitem{Li17-ALF}
J.~Li, W.~Monroe, T.~Shi, S.~Jean, A.~Ritter, and D.~Jurafsky, ``Adversarial
  learning for neural dialogue generation,'' in \emph{Proc.\ Conference on
  Empirical Methods in Natural Language Processing (EMNLP)}, Copenhagen,
  Denmark, 2017, pp. 2157--2169.

\bibitem{Ding18-EFE}
H.~Ding, K.~Sricharan, and R.~Chellappa, ``{ExprGAN: F}acial expression editing
  with controllable expression intensity,'' in \emph{Proc.\ 32nd Conference on
  Association for the Advancement of Artificial Intelligence (AAAI)}, New
  Orleans, LA, 2018, pp. 6781--6788.

\bibitem{Zhou17-PFE}
Y.~Zhou and B.~E. Shi, ``Photorealistic facial expression synthesis by the
  conditional difference adversarial autoencoder,'' in \emph{Proc.\ 7th
  International Conference on Affective Computing and Intelligent Interaction
  (ACII)}, San Antonio, TX, 2017, pp. 370--376.

\bibitem{Song17-GGA}
L.~Song, Z.~Lu, R.~He, Z.~Sun, and T.~Tan, ``Geometry guided adversarial facial
  expression synthesis,'' \emph{arXiv preprint arXiv:1712.03474}, Dec. 2017.

\bibitem{Hsu17-VCF}
C.~Hsu, H.~Hwang, Y.~Wu, Y.~Tsao, and H.~Wang, ``Voice conversion from
  unaligned corpora using variational autoencoding {Wasserstein} generative
  adversarial networks,'' in \emph{Proc.\ 18th Annual Conference of the
  International Speech Communication Association (INTERSPEECH)}, Stockholm,
  Sweden, 2017, pp. 3364--3368.

\bibitem{Perarnau16-ICG}
G.~Perarnau, J.~van~de Weijer, B.~Raducanu, and J.~M. {\'A}lvarez, ``Invertible
  conditional gans for image editing,'' in \emph{Proc.\ Adversarial Training
  Workshop in NIPS}, Barcelona, Spain, 2016.

\bibitem{Lee18-SSF}
C.~Lee, Y.~Wang, T.~Hsu, K.~Chen, H.~Lee, and L.~Lee, ``Scalable sentiment for
  sequence-to-sequence chatbot response with performance analysis,'' in
  \emph{Proc.\ IEEE International Conference on Acoustics, Speech and Signal
  Processing (ICASSP)}, Calgary, Canada, 2018, pp. 6164--6168.

\bibitem{Schuller12-SSF}
B.~Schuller, Z.~Zhang, F.~Weninger, and F.~Burkhardt, ``Synthesized speech for
  model training in cross-corpus recognition of human emotion,''
  \emph{International Journal of Speech Technology, Special Issue on New and
  Improved Advances in Speaker Recognition Technologies}, vol.~15, no.~3, pp.
  313--323, Sep. 2012.

\bibitem{Zhang17-AAN}
Y.~Zhang, R.~Barzilay, and T.~S. Jaakkola, ``Aspect-augmented adversarial
  networks for domain adaptation,'' \emph{Transactions of the Association for
  Computational Linguistics}, vol.~5, pp. 515--528, Dec. 2017.

\bibitem{Zhu17-ECW}
X.~Zhu, Y.~Liu, J.~Li, T.~Wan, and Z.~Qin, ``Emotion classification with data
  augmentation using generative adversarial networks,'' in \emph{Proc.\
  Pacific-Asia Conference on Knowledge Discovery and Data Mining (PAKDD)},
  Melbourne, Australia, 2018, pp. 349--360.

\bibitem{Sahu17-AAF}
S.~Sahu, R.~Gupta, G.~Sivaraman, W.~AbdAlmageed, and C.~Y. Espy{-}Wilson,
  ``Adversarial auto-encoders for speech based emotion recognition,'' in
  \emph{Proc.\ 18th Annual Conference of the International Speech Communication
  Association (INTERSPEECH)}, Stockholm, Sweden, 2017, pp. 1243--1247.

\bibitem{Ganin16-DAT}
Y.~Ganin, E.~Ustinova, H.~Ajakan, P.~Germain, H.~Larochelle, F.~Laviolette,
  M.~Marchand, and V.~S. Lempitsky, ``Domain-adversarial training of neural
  networks,'' \emph{Journal of Machine Learning Research}, vol.~17, no.~1, pp.
  2096--2030, Jan. 2016.

\bibitem{Li17-EAM}
Z.~Li, Y.~Zhang, Y.~Wei, Y.~Wu, and Q.~Yang, ``End-to-end adversarial memory
  network for cross-domain sentiment classification,'' in \emph{Proc.\ 26th
  International Joint Conference on Artificial Intelligence (IJCAI)},
  Melbourne, Australia, 2017, pp. 2237--2243.

\bibitem{Shen18-WDG}
J.~Shen, Y.~Qu, W.~Zhang, and Y.~Yu, ``Wasserstein distance guided
  representation learning for domain adaptation,'' in \emph{Proc.\ 32nd
  Conference on Association for the Advancement of Artificial Intelligence
  (AAAI)}, New Orleans, LA, 2018, pp. 4058--4065.

\bibitem{Zhao18-MSD}
H.~Zhao, S.~Zhang, G.~Wu, J.~Costeira, J.~Moura, and G.~Gordon, ``Multiple
  source domain adaptation with adversarial learning,'' in \emph{Proc.\
  Workshop in 6th International Conference on Learning Representations (ICLR)},
  Vancouver, Canada, 2018.

\bibitem{Liu18-AML}
P.~Liu, X.~Qiu, and X.~Huang, ``Adversarial multi-task learning for text
  classification,'' in \emph{Proc.\ 55th Annual Meeting of the Association for
  Computational Linguistics (ACL)}, Vancouver, Canada, 2017, pp. 1--10.

\bibitem{Chen18-MAN}
X.~Chen and C.~Cardie, ``Multinomial adversarial networks for multi-domain text
  classification,'' in \emph{Proc.\ 16th Annual Conference of the North
  American Chapter of the Association for Computational Linguistics (NAACL)},
  New Orleans, LA, 2018, pp. 1226--1240.

\bibitem{goodfellow2015explaining}
I.~Goodfellow, J.~Shlens, and C.~Szegedy, ``Explaining and harnessing
  adversarial examples,'' in \emph{Proc.\ 3rd International Conference on
  Learning Representations (ICLR)}, Vancouver, Canada, 2015.

\bibitem{Chang17-LRO}
J.~Chang and S.~Scherer, ``Learning representations of emotional speech with
  deep convolutional generative adversarial networks,'' in \emph{Proc.\ IEEE
  International Conference on Acoustics, Speech and Signal Processing
  (ICASSP)}, New Orleans, LA, 2017, pp. 2746--2750.

\bibitem{miyato2016distributional}
T.~Miyato, S.-i. Maeda, M.~Koyama, K.~Nakae, and S.~Ishii, ``Distributional
  smoothing with virtual adversarial training,'' in \emph{Proc.\ 4th
  International Conference on Learning Representations (ICLR)}, San Juan, PR,
  2016.

\bibitem{Miyato17-ATM}
T.~Miyato, A.~M. Dai, and I.~Goodfellow, ``Adversarial training methods for
  semi-supervised text classification,'' in \emph{Proc.\ 5th International
  Conference on Learning Representations (ICLR)}, Toulon, France, 2017.

\bibitem{Sahu18-SMP}
S.~Sahu, R.~Gupta, G.~Sivaraman, and C.~Espy-Wilson, ``Smoothing model
  predictions using adversarial training procedures for speech based emotion
  recognition,'' in \emph{Proc.\ IEEE International Conference on Acoustics,
  Speech and Signal Processing (ICASSP)}, Calgary, Canada, 2018, pp.
  4934--4938.

\bibitem{Arora17-GAE}
S.~Arora, R.~Ge, Y.~Liang, T.~Ma, and Y.~Zhang, ``Generalization and
  equilibrium in {Generative Adversarial Nets (GANs)},'' in \emph{Proc.\ 34th
  International Conference on Machine Learning (ICML)}, Sydney, Australia,
  2017, pp. 224--232.

\bibitem{Salimans16-ITF}
T.~Salimans, I.~Goodfellow, W.~Zaremba, V.~Cheung, A.~Radford, and X.~Chen,
  ``Improved techniques for training {GANs},'' in \emph{Proc.\ Advances in
  Neural Information Processing Systems (NIPS)}, Barcelona, Spain, 2016, pp.
  2226--2234.

\bibitem{Metz16-UGA}
L.~Metz, B.~Poole, D.~Pfau, and J.~Sohl{-}Dickstein, ``Unrolled generative
  adversarial networks,'' in \emph{Proc.\ 5th International Conference on
  Learning Representations (ICLR)}, Toulon, France, 2017.

\bibitem{Tolstikhin17-ABG}
I.~O. Tolstikhin, S.~Gelly, O.~Bousquet, C.~Simon{-}Gabriel, and
  B.~Sch{\"{o}}lkopf, ``{AdaGAN: Boosting} generative models,'' in \emph{Proc.\
  Advances in Neural Information Processing Systems (NIPS)}, Long Beach, CA,
  2017, pp. 5430--5439.

\bibitem{Yang17-SPS}
S.~Yang, L.~Xie, X.~Chen, X.~Lou, X.~Zhu, D.~Huang, and H.~Li, ``Statistical
  parametric speech synthesis using generative adversarial networks under a
  multi-task learning framework,'' in \emph{Proc.\ IEEE Automatic Speech
  Recognition and Understanding Workshop (ASRU)}, Okinawa, Japan, 2017, pp.
  685--691.

\bibitem{Dunbar97-HCB}
R.~I. Dunbar, A.~Marriott, and N.~D. Duncan, ``Human conversational behavior,''
  \emph{Human Nature}, vol.~8, no.~3, pp. 231--246, Sep. 1997.

\end{thebibliography}

\end{document}